# Adversarial Machine Learning for Social Good: Reframing the Adversary as an Ally

*Shawqi Al-Maliki*[*], *Adnan Qayyum*[†], *Hassan Ali*[†], *Mohamed Abdallah*[*], *Senior Member, IEEE, Junaid Qadir*[‡], *Senior Member, IEEE, Dinh Thai Hoang*[§], *Senior Member, IEEE, Dusit Niyato*[¶], *Fellow, IEEE, Ala Al-Fuqaha*[*], *Senior Member, IEEE*

[*] Information and Computing Technology (ICT) Division, College of Science and Engineering, Hamad Bin Khalifa University, Doha 34110, Qatar
[†] Information Technology University, Lahore, Pakistan
[‡] Department of Computer Science and Engineering, College of Engineering, Qatar University, Doha, Qatar
[§] School of Electrical and Data Engineering, University of Technology Sydney, Australia
[¶] School of Computer Science and Engineering, Nanyang Technological University, Singapore

*Abstract*—Deep Neural Networks (DNNs) have been the driving force behind many of the recent advances in machine learning. However, research has shown that DNNs are vulnerable to adversarial examples—input samples that have been perturbed to force DNN-based models to make errors. As a result, Adversarial Machine Learning (AdvML) has gained a lot of attention, and researchers have investigated these vulnerabilities in various settings and modalities. In addition, DNNs have also been found to incorporate embedded bias and often produce unexplainable predictions, which can result in anti-social AI applications. The emergence of new AI technologies that leverage Large Language Models (LLMs), such as ChatGPT and GPT-4, increases the risk of producing anti-social applications at scale. AdvML for Social Good (AdvML4G) is an emerging field that repurposes the AdvML bug to invent pro-social applications. Regulators, practitioners, and researchers should collaborate to encourage the development of pro-social applications and hinder the development of anti-social ones. In this work, we provide the first comprehensive review of the emerging field of AdvML4G. This paper encompasses a taxonomy that highlights the emergence of AdvML4G, a discussion of the differences and similarities between AdvML4G and AdvML, a taxonomy covering social good-related concepts and aspects, an exploration of the motivations behind the emergence of AdvML4G at the intersection of ML4G and AdvML, and an extensive summary of the works that utilize AdvML4G as an auxiliary tool for innovating pro-social applications. Finally, we elaborate upon various challenges and open research issues that require significant attention from the research community.

## I. INTRODUCTION

In recent years, there has been a significant focus on the vulnerability of deep learning-based models to adversarially perturbed inputs [1]. Specifically, Deep Neural Networks (DNNs) have been shown to be susceptible to various types of adversarial attacks, such as evasion attacks [1]. The ML and computer security communities have collaborated to address these vulnerabilities and develop effective countermeasures, leading to the field of *Adversarial ML* (AdvML), which focuses on attacks and countermeasures to enhance the robustness of DNNs against harmful attacks [2]–[4]. Despite the considerable efforts invested by researchers in AdvML, the challenge of strengthening DNNs against adversarial attacks remains unresolved due to inherent limitations in how DNNs work. Moreover, leading experts in the AdvML community are not optimistic about this challenge being resolved in the near future [5]–[8]. On the other hand, the AdvML problem has not posed a significant threat to real-world complex ML systems beyond the narrow and isolated problems studied in research laboratories. These limitations, along with others, weaken the motivation of AdvML and raise questions about the feasibility of further research efforts in such a direction. If adversarial attacks are less effective beyond academic laboratories, these attacks might not matter!

In current AdvML research, the main focus is on reducing negative technical impacts on DNN performance by enhancing their adversarial robustness through proposing attacks and defenses, playing the role of red and blue teaming (as illustrated in Fig. 2). Therefore, current AdvML research is associated with the harmful connotation of protecting ML models against potential real adversaries. However, adversarial attacks should be considered neutral tools. They can be utilized to convey both positive and harmful connotations. While the proposed attacks are neutral and could have a positive social impact within this context, the majority of AdvML works do not prioritize the development of socially beneficial ML applications. Even when researchers engage in "*red teaming*" practices, involving the attack and evaluation of systems, they typically assume the role of a potential real adversary to assess and reinforce system resilience against such risks. However, it is important to acknowledge that adversarial attacks should be regarded as neutral, and their effects depend on the context and underlying intent. For instance, they can be employed to counter the use of ML-based systems by authoritarian oppressive governments that infringe upon human privacy, such as face recognition and emotion recognition systems.

The emergence of new AI technologies, such as GPT-4 and tools like ChatGPT, raises the concern of an increased risk in the development of anti-social applications. To address this, collaboration among regulators, practitioners, and researchers becomes crucial in promoting pro-social applications and preventing the proliferation of socially problematic ones. Researchers in the AdvML community have a significant role in this effort by repurposing AdvML attacks for socially beneficial applications, thus establishing and advancing the field of "*AdvML for Social Good*" (AdvML4G).

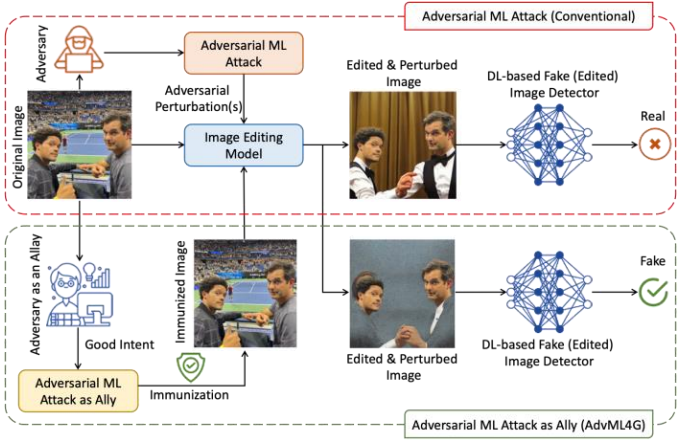

Fig. 1: Illustrating an example of a DNN-based fake image detector for two cases: (1) edited image is perturbed by the adversary (using AdvML) to evade detection (top), and (2) edited and perturbed image is immunized by leveraging AdvML as an ally (AdvML4G) to make fake image detector resilient (bottom). Individual images taken from [10].

Relying on the definition of AI for social good [9], we define AdvML4G as "*the emerging research direction that involves innovating ML-based systems that are developed and deployed based on the learned lessons from the traditional AdvML (i.e., adversarial robustness) to: (i) prevent, mitigate and/or resolve problems adversely impacting human life, and/or (ii) enable socially beneficial applications, while (iii) not introducing new kinds of societal harm.*" We use the term AdvML4G to refer to any adversarial attack motivated by legitimate human rights or humanitarian social good. AdvML4G is a special type of "*ML for Social Good*" (ML4G), where AdvML enables the "social good" aspect, while ML4G encompasses any ML enablers.

AdvML4G aims to go beyond the sole technical aspect of AdvML to embrace a socio-technical aspect where both technical and social considerations matter, with greater emphasis on the societal dimension. In the context of AdvML4G, the term "adversary" adopts a different meaning, acting more like an ally. Unlike traditional AdvML research, AdvML4G aims to extend beyond adversarial robustness to innovate applications that are beneficial for society. Instead of using attacks solely to identify and mitigate model limitations, AdvML4G leverages insights gained from AdvML research to develop applications striving for social good. As a result, AdvML4G attacks differ significantly from typical attacks crafted by real or virtual adversaries. Moreover, AdvML4G strives to ensure the sustainability of the field by emphasizing the importance of social good. This emerging direction encourages AdvML research to continue to expand while prioritizing societal benefits.

Fig. 1 provides an example that demonstrates the distinction between the pipelines of AdvML and AdvML4G. The top subfigure illustrates how AdvML researchers simulate adversarial attacks against DNN-based fake image detectors to enhance their adversarial robustness. The simulated adversarial attacks enable the edited image to evade the fake image detection. On the other hand, the bottom subfigure illustrates how AdvML4G researchers (adversaries as allies) leverage AdvML for social good. In particular, it illustrates how AdvML attacks can be used to immunize user images before sharing them online, thereby mitigating misuse by enabling fake image detectors to identify them as fake.

To the best of our knowledge, this is the first work that provides a comprehensive review of AdvML4G as an emerging area, including introducing the emergence and motivation of AdvML4G; the evolution of the field of AdvML towards AdvML4G; and an extensive summary of the works that consider AdvML as a tool for innovating socially beneficial applications. While some existing works have acknowledged the AdvML4G phenomenon [11], [12], none of them offers a comprehensive survey of relevant published applications. Albert et al. [11] assesses societal values within the AdvML community, exploring whether published adversarial ML works consider the broader impact on society. The study focuses on papers presented at NeurIPS (2020), where the authors were requested to address the positive and negative social impact of their work. However, it does not provide a summary of works specifically related to AdvML4G. Similarly, Chen et al. [12] presented some applications resulting from insights gained from adversarial robustness but do not introduce AdvML4G as an emerging research direction nor provide a comprehensive summary of AdvML4G applications.

The sophisticated variations of popular AdvML attacks, such as evasion [1], poisoning [13], and reprogramming attacks [14], are primarily introduced and discussed in the context of DNN models due to their remarkable performance compared to traditional ML models. Therefore, in this paper, we focus on DNN-based AdvML4G applications.

Before moving further, it is important to highlight that this survey is comprehensive but cannot be exhaustive for two reasons. Firstly, some socially beneficial applications are published without explicitly highlighting their socially good aspects. For instance, they don't mention social good concepts or aspects outlined and discussed in Section II-B2. Secondly, AdvML4G is still a nascent field, which means that numerous social aspects remain unexplored within it.

The salient contributions of this work are summarized as follows:

1) We introduce a taxonomy to highlight the emergence of AdvM4G at the intersection of AdvML and *ML4G* research domains and discuss the evolution of the AdvML field (Section II-A).
2) We highlight the differences and similarities between AdvML4G and AdvML (Section II-B1), and provide a taxonomy for the *social good* related concepts and aspects (Section II-B2).
3) We shed light on the motivations behind the emergence of AdvML4G (Section III).
4) We provide the first comprehensive review of the emerging field of AdvML4G, where AdvML is leveraged

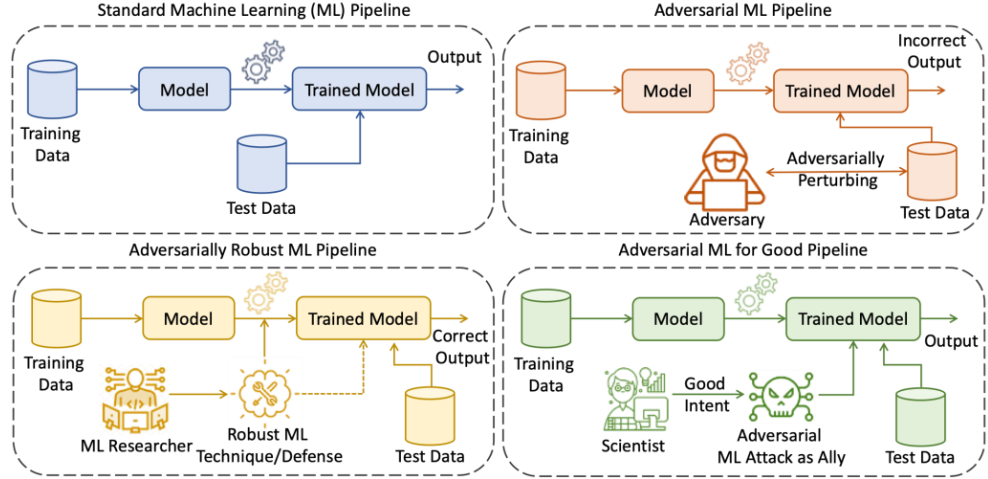

Fig. 2: ML pipelines for: (1) Standard ML; (2) Adversarial ML; (3) Adversarially Robust ML; and (4) AdvML4G. **Standard ML** pipeline shows a non-security-critical scenario. The **Adversarial ML** pipeline represents red teaming, where AdvML researchers simulate real adversaries to identify vulnerabilities. **Adversarially Robust ML** pipeline represents blue teaming, where AdvML researchers propose defenses against the identified vulnerabilities. **AdvML4G** pipeline demonstrates how adversaries, acting as allies, can utilize AdvML to mitigate the negative impact of socially harmful deployed models.

as a tool for producing socially beneficial applications (Section IV).
5) We highlight challenges faced by researchers in AdvML4G (Section V), outline future research directions (Section VI), and provide recommendations on the roles that governments, industry, and academia can play to promote AdvML4G (Section VII).

We organize the remaining of this paper as follows. Section II presents the emergence of AdvML4G and introduces the fundamental background. In Section III, we detail the reasons beyond moving AdvML towards ML4G. In Section IV, we show a detailed summary of the AdvML4G applications. Section V delineates the challenges of AdvML4G while future research directions are highlighted in Section VI. Finally, we introduce the recommended roles that governments, industry, and academia should play to promote AdvML4G in SectionVII and conclude the paper in Section VIII.

## II. BACKGROUND AND EMERGENCE OF ADVML4G

This section introduces the emergence of AdvML4G and explains the fundamental concepts that facilitate understanding this paper.

### A. Emergence of AdvML4G

This section presents the position of AdvML4G amongst ML4G applications and AdvML aspects. Specifically, in Section II-A1, we show how AdvML4G is positioned within ML4G applications, and in Section 4, we demonstrate how it is positioned and evolved within AdvML aspects. Fig. 3 illustrates the emergence of AdvML4G at the intersection of AdvML and ML4G applications.

*1) ML4G Applications:* Conventional *ML4G* applications are developed to enable the social good outcome from the early stages of application development utilizing conventional ML tools and following best "ML4G" practices throughout the model development pipeline. For example, they avoid using biased, private, or harmful data during data collection, incorporate model explainability techniques during training, and take into account ethics and human rights after model deployment (during inference time). The owners and stakeholders of such applications show responsibility towards society and embrace enabling social good outcome. However, they may not reach the full potential of creating socially beneficial applications due to the constraints inherent in conventional ML development tools. To close this gap, AdvML attacks can be utilized as allies to extend the development of ML4G applications to new ones that cannot be innovated otherwise (i.e., AdvML4G applications) as detailed below.

In AdvML4G applications, enabling the social good outcome can be promoted either by the application owners and stakeholders during application development to innovate socially good capabilities, or it can be enabled by the potentially affected entities while the model is serving (i.e., after application deployment) to mitigate socially harmful aspects. For innovating AdvML4G applications, the model owner utilizes AdvML attacks as allies to enable capabilities that are challenging to be developed with conventional ML techniques. For example, the model owner can utilize backdoor attacks (type of poisoning attacks explained in Section II-B6) to inject a watermark into the model to protect the owner against intellectual property infringements. Therefore, the owner can claim the model ownership at any time at inference time by embedding any input with a specific pattern (called a trigger) that a model learned during the training time.

On the other hand, potentially affected entities can utilize AdvML attacks as an auxiliary tool to enable the social good outcome into applications deployed with potential social harms. For example, a work proposed by Salman et al. [10] utilizes adversarial examples to protect users' privacy by mitigating the misuse of diffusion models to generate harmful images based on a given image. In other words, they enable immunizing user images before sharing them online, which renders the shared images resistant to manipulation by diffusion models. This immunization is achieved by injecting imperceptible adversarial perturbations that disrupt the processes of the targeted diffusion models, which force them to generate unrealistic images (further details in Section IV-C). Section IV provides detailed examples of utilizing AdvML4G in innovating socially good capabilities or mitigating socially harmful ones.

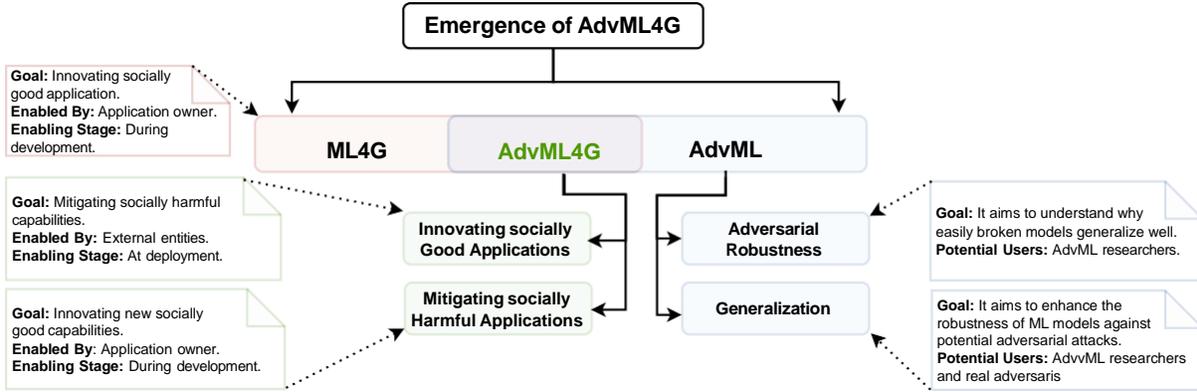

Fig. 3: Illustration of the emergence of AdvML4G at the intersection of AdvML and ML4G, highlighting the root of AdvML4G. ML4G can be extended to include applications enabled by AdvML attacks as allies. Similarly, AdvML attacks can be extended to include attacks acting as allies to enable developing a special type of ML4G applications (i.e. AdvML4G applications) that cannot be developed otherwise.

*2) AdvML Aspects and Evolution:* AdvML (DNNs-based) emerged in 2013 with two parallel research directions. One direction tackles the generalization aspect, while another tackles the security aspect. The early works such as [1], [15], [16] focus on the generalization aspect of AdvML. They aim to understand what model is learning and why easily broken models tend to generalize well. In other words, if models are so brittle in adversarial settings, why do they generalize (predict well) in benign settings? Do they learn meaningful patterns or just spurious correlations?

Another direction addressing AdvML as a security concern (i.e., adversarial robustness) is exemplified by the work of Biggio et al. [17], Kurakin et al. [18], Papernot et al. [19], and Madry et al. [2]. They tried to identify and mitigate the limits of the ML models by inventing attacks and developing corresponding defenses. This line of work aims to enhance the robustness of ML models against potential adversarial attacks. It has gained tremendous success in academic research (i.e., dominating AdvML research) but encountered challenges to be applied in real systems (as detailed in Section III-B). Research on advML is dominated by the adversarial robustness aspect. Thus, in this work, we use the terms "adversarial robustness", "Traditional AdvML", and "AdvML" interchangeably.

AdvML attacks acting as allies contribute to innovating a special variation of ML4G known as AdvML4G. AdvML4G

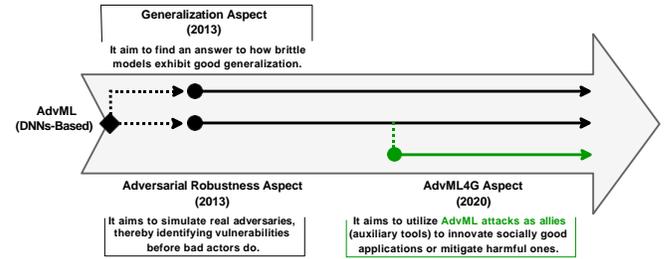

Fig. 4: Evolution of AdvML (DNNs-based) research. Researchers have acknowledged that while some adversarial failures can be addressed, not all can be, and the invented attacks may not pose a real-world threat. This has led to the emergence of AdvML4G.

has emerged to sustain the growth of AdvML research and utilize the achieved progress toward pro-social ML applications. It is an emerging line of work beyond the adversarial robustness aspect. AdvML4G aims to repurpose the adversarial robustness to be utilized for innovating socially beneficial applications. It has been adopted in various domains (as explained in Section IV), and it is expected to keep growing within those domains and extending to further domains and applications.

AdvML attacks have a different role in each aspect. AdvML researchers employ AdvML attacks within the generalization aspect to find an answer to how brittle models can still exhibit good generalization. In the aspect of adversarial robustness, researchers utilize AdvML attacks to simulate real adversaries, thereby identifying vulnerabilities before bad actors do. Conversely, within the AdvML4G context, researchers leverage AdvML attacks as tools to innovate socially good applications or mitigate socially harmful ones. Table I provides descriptions for AdvML aspects and presents representative works corresponding to each aspect, while Fig. 4 depicts the evolution of these AdvML aspects.

TABLE I: Representative works in AdvML research (focused on DNNs).

| # | AdvML Research Stage | Description | Sample Works | Stage's Advent |
|---|---|---|---|---|
| 1 | AdvML as a generalization concern | This aspect highlights that the existence of the adversarial examples is in contradiction with the capability of DNNs to obtain high generalization performance. If DNNs can generalize well, how can the imperceptible perturbations confuse them? | Szegedy et al. [1], Goodfellow et al. [15] | 2013 |
| 2 | AdvML as a security concern (Adversarial Robustness) | The adversarial robustness refers to a model's ability to resist being fooled. It aims to robustify DNN models by proposing attacks to identify the limitations and evaluate the proposed defenses. | Biggio et al. [17], Kurakin et al. [18], Papernot et al. [19], Madry et al. [2] | 2013 |
| 3 | AdvML4G | The emerging research direction that involves innovating ML-based systems that are developed and deployed based on the learned lessons from adversarial robustness to: (i) prevent, mitigate, resolve problems adversely impacting human life; (ii) enable socially beneficial applications, while (iii) not introducing new kinds of societal harm. | Kumar et al. [20] Salman et al. [10] Albert et al. [11] Chen et al. [21] | 2020 |

*B. Background*

*1) Comparing AdvML and AdvML4G: Goals, Similarities, and Differences:* This section demonstrates a comparison between AdvML and AdvML4G. We begin by illustrating AdvML and AdvML4G alongside other related ML pipelines (Fig. 2). Then, we elaborate on the comparison, considering ultimate goals, similarities, and differences. After that, the comparison summary is presented in Table II.

**ML Pipelines:** To make the distinction between AdvML and AdvML4G even clearer, Fig. 2 demonstrates them alongside relevant pipelines: 1) The Standard ML pipeline shows a non-security-critical scenario where the existence of an adversary is not assumed. Thus, predictions are often correct (assuming a well-trained model). 2) The Adversarial ML pipeline represents red teaming, where AdvML researchers simulate real adversaries to identify vulnerabilities. 3) The Adversarially Robust ML pipeline represents blue teaming, where AdvML researchers propose defenses against the detected vulnerabilities to make the model robust against potential real-world adversarial attacks. 4) The AdvML4G pipeline demonstrates how adversaries, acting as allies, can utilize AdvML attacks to mitigate the negative impact of socially harmful deployed models by aligning the model's output with social good outcomes.

**AdvML: Goal and Strategy.** The ultimate goal of AdvML research is to render safety and security-critical applications robust (attack-free) against all potential attacks to ensure reliable predictions. To attain this goal, the AdvML community adopted a long-run strategy. In particular, due to the lack of publicly published end-to-end ML systems [22] and the challenges in their threat model, researchers in AdvML have started tackling the robustness issue by considering bare ML models rather than ML systems with the hope that the research progress would end up with sophisticated solutions that can be scaled and applied to end-to-end ML systems.

In addition, they have started publishing works that assume constrained adversaries (i.e., unrealistic threat model) with the hope that the accumulative research work would result in robust solutions against unconstrained adversaries, i.e., realistic threat model (further details on the limitations of AdvML are on Section III-A). However, none of these hopes has become real after a decade of massive work. They have achieved limited outcomes on real systems as explained in Section III-B. These limited research outcomes contribute to accelerating the uncovering of AdvML4G as an overlooked yet essential research direction.

**AdvML4G: Goal and Strategy.** The ultimate goal of AdvML4G is to utilize adversarial ML attacks as an auxiliary tool to innovate socially good applications or mitigate socially harmful ones whenever conventional ML tools cannot help. In addition, AdvML4G broadens the scope of AdvML and sustains its growth as an active area of research. Researchers have been approaching this goal by leveraging the learned lessons and the proposed attacks in AdvML and enabling the social good outcome in various applications (as detailed in Section IV). Because of the limitations in AdvML research (as detailed in Section III-A), some researchers started questioning the feasibility of pursuing further research in that direction. Moreover, they have noticed they neglected a substantial socio-technical aspect beyond robustifying ML systems. Specifically, they have overlooked utilizing the proposed attacks in innovating (not robustifying) ML4G applications that cannot be innovated otherwise. AdvML attacks shine in innovating externally enabled pro-social applications where conventional ML development techniques are unable to bring up the social good outcome (see Section II-A).

**Similarities and Differences** AdvML and AdvML4G are similar in the sense that both utilize proposed AdvML attacks but for different purposes. The underlying technology for AdvML and AdvML4G is the same. AdvML utilizes AdvML attacks to enhance security, while AdvML4G utilizes them as allies to innovate socially good capabilities. On the other hand, AdvML and AdvML4G have many distinctions. AdvML has an implicit aspect of social good, which is safety. However, this aspect has not been explicitly highlighted as a social good dimension in the relevant literature. The focus of AdvML lies on the sole technical aspect (i.e., defending against adversarial attacks) rather than the socio-technical aspects encompassed in AdvML4G. To highlight the distinction between AdvML and AdvML4G, we compare them with respect to their goals, applications, and threat model, as outlined below.

- Goals: AdvML focuses mainly on the security (i.e., adversarial robustness) of ML models. AdvML4G goes beyond the adversarial robustness to include all social good perspectives, including safety and robustness. In other words, AdvML addresses the technical aspect, whereas

TABLE II: Comparison of AdvML and AdvML4G in Terms of Goals, Similarities, and Differences.

|  |  | AdvML | AdvML4G |
|---|---|---|---|
| **Ultimate Goals** | | Rendering safety and security-critical applications robust (attack-free) against all potential attacks to ensure reliable predictions. | 1) Utilize adversarial ML attacks as an auxiliary tool to innovate socially good applications or mitigate socially harmful ones whenever conventional tools cannot help. 2) AdvML4G broadens the scope of AdvML and sustains its growth as an active area of research |
| **Similarities** | | 1) The underlying technology for both (AdvML and AdvML4G) is the same. AdvML utilizes AdvML attacks to enhance security, while AdvML4G utilizes them as allies to innovate socially good capabilities. 2) Since safety and security applications are part of the applications covered by AdvML4G, both AdvML and AdvML4G share the same threat model for this set of applications. | |
| **Differences** | Goal | It focuses mainly on the security (i.e., adversarial robustness) of ML models | It goes beyond the adversarial robustness to include all social good perspectives, including safety, and robustness. |
| | Applications | AdvML is considered in scenarios where safety and security are concerns. | AdvML4G encompasses AdvML applications. It covers broader domains than AdvML. In particular, it extends safety and security-critical applications to include *social good* aspects such as privacy, fairness, and innovation |
| | Threat Model | AdvML assumes the existence of attackers and defenders. | AdvML4G doesn't assume the existence of defenders. |

AdvML4G encompasses the socio-technical dimensions.
- Applications: AdvML is considered in scenarios where safety and security are concerns. On the other hand, AdvML4G encompasses AdvML applications and covers broader domains. In particular, it extends safety and security-critical applications to include *social good* aspects such as privacy, fairness, and innovation.
- Threat model: As safety and security applications are covered by both AdvML4G and AdvML. For this subset of applications, AdvML4G and AdvML share the same threat model. Other than that, AdvML4G adopts a distinct threat model. Specifically, it doesn't assume the existence of defenders.

*2) ML4G Related Concepts and Potential Outcome:* ML for Social good (ML4G goes in the ML literature with different concepts, such as *ML Safety, Pro-Social ML, Human-Centered ML, Socially Good ML, Socially Beneficial ML, ML Alignment, Trustworthy ML, Responsible ML, ML Ethics*. The body of literature refers to these concepts to imply one or more ML4G potential outcome. They are overloaded and loosely used concepts and there is no clear distinction between their definitions. Depending on the context, they may refer to a different spectrum of ML4G potential outcomes. By *ML4G potential outcome*, we mean the potential positive societal outcome result from applying ML to address societal challenges, including ML ethics. Main examples of ML4G potential outcome are *Honesty, Inclusion, Transparency, Innovation, Harmlessness, Privacy, Helpfulness, Justice, Equity, Reliability, Robustness, Accountability, Fairness*. Fig. 5 depicts the idea of how ML4G might refer to more or less ML4G potential outcome. £The connected arrows in Fig. 5 highlight that these concepts are overlapping and loosely used to imply that a ML4G concept may encompass one or multiple ML4G potential outcomes. In this work, we choose to use *socially good* and *pro-social* ML concepts interchangeably as representatives for various ML4G concepts, due to their popularity in AdvML-related literature. The *ML4G* related concepts and potential outcome are self-explanatory and all have the theme and connotations of societal benefits. As this work reviews a narrower field of *ML4G*, which is AdvML4G, going deeper with the description and examples of all the related concepts and their corresponding ML4G potential outcome is out of the scope of this work.

When a social good outcome results from utilizing ML, we refer to the outcome as ML4G application (detailed in Section II-A1). Likewise, when the social good outcome results from utilizing AdvML, we refer to the corresponding outcome as AdvML4G application (detailed in Section IV). The focus of this review work is AdvML4G. Since this is an emerging area of research, there are no associated applications for all ML4G potential outcome. In addition, some AdvML4G applications do not use explicit *ML4G* concepts, which makes finding these works challenging. Therefore, this survey is comprehensive but not exhaustive (as we highlighted in the Introduction Section).

*3) Anti-Social AdvML:* One may ask: since there is pro-social AdvML, what is anti-social AdvML, then? This valid question can be addressed by presenting two anti-social cases. The first case is when a real adversary leverages publicly published state-of-the-art adversarial attacks to harm pro-social applications, such as evading authorized face recognition systems or bypassing spam detection systems. The second case is when real adversaries develop anti-social applications. An example of such anti-social applications is the face recognition system that was developed by Clearview company. Clearview scrapes publicly available images and builds an application that violates human privacy by enabling the identification of the face of any person [23]. The ultimate goal of *adversarial robustness* aspect of AdvML is to prevent such anti-social potentials. AdvML attack is a double-edged sword, in which a real adversary can utilize it to offend society while an adversary as-ally can utilize it to help society.

*4) Threat Model:* A threat model is defined by the goals, capabilities, and knowledge the adversary is assumed to utilize for crafting an adversarial attack [24], [25]. Black-box and white-box are the most popular threat models. White-box threat model assumes that the adversary has full knowledge of the model's internal information as well as the output. The adversary knows the model architecture, its parameters, and the outputs of the models. In the black-box threat model, the adversary has no information about the model architecture or its parameters but has access to the output. The black-box threat model comes in three variants based on the adversary's degree of access to the output: limited query [26], hard-label [27], and soft-label [28] black-box threat models. In the limited-query variant, the allowed number of submitted queries to the model is limited. For the soft variant, the output of the query is score-based (probability prediction), while the hard-based variant is label based (without showing the score).

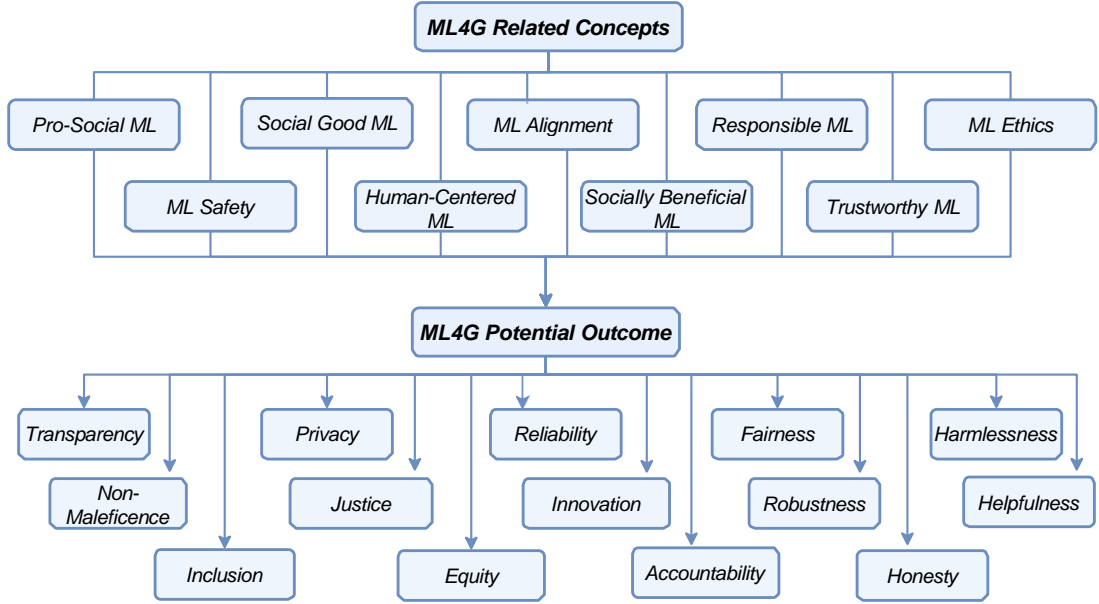

Fig. 5: A taxonomy illustrating *ML4G* related concepts and their corresponding potential outcome. *ML4G* concepts refer to various popular terms that are loosely used to imply one or more ML4G potential outcome. For example, in one context, trustworthy ML (as a ML4G concept) may refer primarily to robustness, while in another context, it may encompass multiple ML4G potential outcomes, such as robustness, privacy, fairness, and transparency.

*5) Distance Metrics:* A distance or similarity metric in AdvML is the measurement that is used for quantifying the differences between the input before and after adding perturbation. $L_p$ norm distance metrics ($L_0$, $L_2$, $L_\infty$) are the most popular similarity metrics (in research) that are used for measuring the adversarial attack. They are useful for theoretical and experimental studies. However, they are unrealistic as they are not aligned with human perception. Another non $L_p$ norm distance metric that moves beyond $L_p$ norms is Wasserstein norm [29]. Wasserstein norm can capture common notions of image transformations such as translations, distortions, and rotation.

*6) Adversarial Attacks:* Researchers contribute to adversarial attacks as a proactive practice to identify the limits of ML models. Adversarial attacks can happen at any phase of ML life cycle, i.e., training or inference.

*Training-Time Attacks:* Poisoning attacks [13], [30]–[33] are the most popular adversarial attacks used during the model's training phase. They aim to contaminate the training dataset so that the trained model fails to generalize. Also, poisoning attacks can degrade the performance of ML-based systems by causing a denial of service (DoS). ML-based systems that rely on the quality of the dataset are sensitive to poisoning attacks. Backdoor attacks (also called trojan attacks) [34]–[37] are another type of training time attacks. They are crafted by embedding a trigger to a subset of training data that are manipulated to have wrong labels [35] [38]. Backdoor triggers are patterns embedded in a subset of inputs during the model training time. This enables the trained model to establish strong correlations between the embedded trigger and an adversary-chosen wrong label. At inference time, whenever the adversary embeds an input with such a trigger, it induces the model to predict the wrongly chosen label. For the inputs without embedded triggers, model predictions remain correct. The ubiquity of pre-trained ML models and open-sourced datasets makes them prone to backdoor attacks [39].

*Inference-Time Attacks:* The literature suggests that the current neural networks are susceptible to a broad range of AdvML attacks at inference time [2], [26], [40]–[42]. For instance, Szegedy et al. [15] observed the presence of adversarial examples in the image classification task where it is possible to change the predicted label of the image by adding a well-designed small amount of perturbation. Adversarial examples are also called evasion attacks [1]. The body of research suggests various types of evasion attack algorithms [18], [40], [43], [44]. Model stealing attacks [45], membership inference attacks [46], and adversarial reprogramming [14] are other popular types of AdvML inference time attacks.

## III. MOTIVATIONS BEHIND THE EMERGENCE OF ADVML4G

Researchers in the AdvML community have contributed massively to the adversarial robustness aspect of DNNs. They have been attempting to endow DNNs with guaranteed protection against adversarial attacks. However, this challenge is still unaddressed. Even worse, distinguished experts in the AdvML community highlight that this challenge cannot be addressed soon [5], [6], [7], [8]. Furthermore, due to the inherent limitations of the ongoing AdvML academic research (details in Section III-A), AdvML attacks do not pose a remarkable threat to the real systems (as illustrated in Section III-B).

TABLE III: Examples of AdvML attacks on real systems. Currently, very few popular real-world systems have been successfully attacked and compromised.

| Reference | AdvML Attack Type | Threat Model | Description |
|---|---|---|---|
| [48] | Poisoning Attack | Black-box | This work proposes a practical adversarial attack that can directly poison 10 well-known datasets, such as Wikipedia. For example, this work shows that, to poison Wikipedia successfully, adversaries need a limited time window to inject malicious inputs. |
| [49] | Evasion Attacks | White-box | This work shows that the existing visual ad-blockers are vulnerable to evasion attacks. The work explained how to construct imperceptible perturbations in a real web page for various types of advertisements to evade the placed ad-blockers so the ads are displayed to the users. |
| [50] | Evasion Attacks | Black-Box | This paper develops adversarial music that can evade copyright detection. The developed adversarial music managed to fool real-world systems, such as the AudioTag music recognition service [51], and YouTube's Content ID system [52]. |
| [53] | Evasion Attacks | Black-box | This work highlights the significance of utilizing a preprocessor to defend against adversarial attacks. It suggests integrating the preprocessor as a component in real-world systems despite the challenges involved. The work proposes a successful adversarial attack by reverse-engineering the preprocessor and demonstrates the superiority of weak preprocessor-aware attacks over strong preprocessor-unaware attacks. |

### A. Limitations of Traditional AdvML Research

Traditional AdvML research has three major limitations that render AdvML a less effective threat to real-world systems and raise a question on the feasibility of further research efforts in this direction.

*Firstly:* AdvML research works attack stand-alone models, while models in practice are incorporated into larger systems [47]. Thus, real adversaries need to attack the whole system pipeline, not just the model. The whole system pipeline consists of the complete layers or filters within the larger system through which the manipulated input must pass before reaching the model. This real setting makes crafting successful adversarial attacks in practice more challenging.

*Secondly:* Studies on the white-box threat model dominate in the literature. There is less research work with the query-based black-box threat model and even less work with the more realistic settings, such as the limited query-based [48] and score-based black-box threat model [27]. However, black-box threat model attacks usually occur in real systems. This difference in the considered threat model demonstrates a clear gap between AdvML academic research and real-world DNNs-based systems.

*Thirdly:* In traditional AdvML research, generated attacks are assumed to have imperceptible perturbations. However, attacks on real ML systems are not necessarily imperceptible as a real adversary is not bound by constraints that limit attacks to imperceptible levels.

These three limitations hinder the successful launch of attacks on real-world ML systems, and there are very limited published works that investigate realistic ML systems. Therefore, these limitations in AdvML research contribute to the acceleration of uncovering an essential yet overlooked positive aspect of AdvML, which involves utilizing AdvML attacks as allies to innovate pro-social applications.

### B. Limited AdvML Attacks on Real Systems

The claim of the limited published work on adversarial attacks on real systems is further reinforced by the minuscule proportion of AdvML research that focuses on real-world settings compared to the volume of published work that does not. Since the early publications of adversarial examples in late 2013 up to 2023, the publications in the field increased almost exponentially [54]. However, the ratio of the works with realistic settings is only roughly 1% (i.e., 60/6000) [6]. This implies that there is a limited impact of AdvML research in real-world settings.

There are other reasons that make developing AdvML attacks on real systems less attractive. For example, there is limited demand from security practitioners to robustify ML-powered systems against AdvML attacks because AdvML attacks are less common in the real world compared to non-ML-based attacks [55]. Thus, security practitioners are not prioritizing incorporating ML defenses into ML-powered systems because real-world evidence shows that actual adversaries utilize simple approaches (rather than AdvML attacks) to attack ML-powered systems [22].

Moreover, practitioners do not take extensively the security of the deployed ML systems because they believe that the accessibility of the deployed systems is under control. On the other hand, they are more concerned about the attacks that are out of their control, such as social engineering attacks [56].

Table III demonstrates examples of AdvML research works [48], [50], [53] that succeeded in attacking real systems. Some of these works are uncommon real-world threat models, such as the work of Tramer et al. [49]. This work has a white-box threat model. In addition, the model is not deployed as part of a bigger remote system. Instead, it is installed on the client side. Other sample works [48], [50], [53] are with the black-box threat model. These are the adversarial attacks that matter as they are the ones that succeed in real systems. However, these types of works are uncommon due to the above-mentioned limitations. The limitations of traditional AdvML research, which lead to a lack of AdvML attacks on real-world applications, contributed to the idea of moving beyond the technical aspect of AdvML, specifically, beyond the adversarial robustness, and into the socio-technical aspect.

### C. Moving Beyond AdvML towards AdvML4G

There is a tendency towards discouraging publishing AdvML attacks and defenses. The latest talks by influential researchers in the field of AdvML question the feasibility of continuing publishing on the adversarial robustness direction

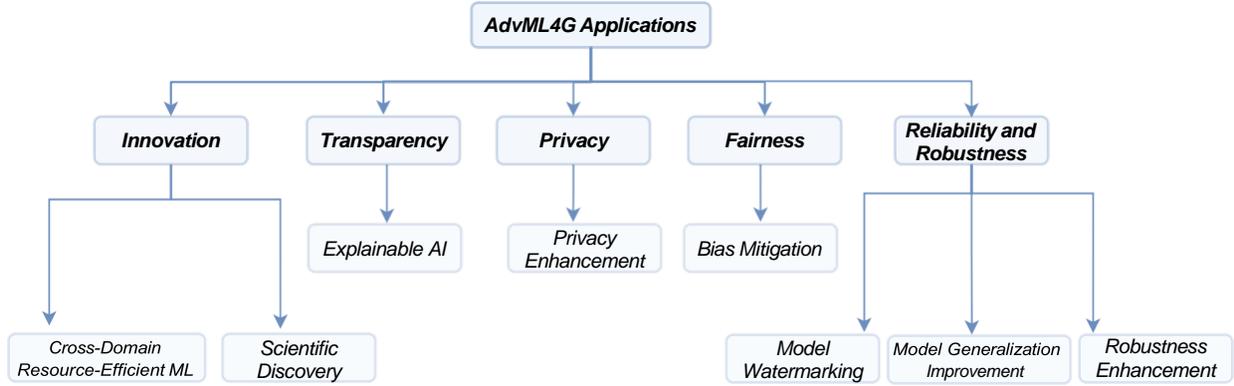

Fig. 6: A taxonomy illustrates the explored AdvML4G applications. While this taxonomy is comprehensive, it may not be exhaustive, as some pro-social applications are published without explicitly highlighting their ML4G potential outcome.

[5]–[7]. The attacks are ineffective in real-world systems, and the defenses lack providing high robustness against the attacks with realistic settings (reasonable perturbation in limited access or score-based black-box threat model). Despite the massive number of published adversarial defenses, the effectiveness of these defenses is limited [57], [58]. However, they contribute to making ML models resilient against some but not all adversarial attacks.

This obvious gap between AdvML in academic research and AdvML in the real world motivates moving towards AdvML4G. More specifically, this gap has raised the question of the feasibility of AdvML research as adversarial robustness in the first place. If real systems have less potential to be exposed to AdvML attacks, why should researchers pay more attention to the adversarial robustness aspect? The efforts of researchers in this field should be utilized appropriately. They should utilize the learned lessons from the AdvML field and explore the potential of applying them to different but related directions, such as innovating pro-social applications. Such an argument gave birth to AdvML4G.

## IV. Applications of AdvML4G

This section presents a summary of various AdvML4G applications. Based on the potential social good outcome results from applying AdvML4G, we categorize the reviewed AdvML4G applications into Innovation, Transparency, Privacy, Fairness, Reliability and Robustness.

### A. Innovation

This subsection reviews AdvML4G innovative applications that lead to advancing the lives of individuals, enhancing their productivity, or facilitating the development of new products that cannot be realized otherwise.

*1) Cross-Domain Resource-Efficient ML:* Adversarial reprogramming [14] has started as a type of attack that contributes to adversarial robustness and then found its way to be utilized for innovating socially good applications. That is, adversarial reprogramming attacks were initially associated with negative connotations as they can be used to consume the resources of online models for the benefit of the adversary. The adversary can utilize online models for a different task than the task instantiated by the provider. Therefore, to enhance the robustness of the online models against such threats, AdvML researchers work on proposing corresponding countermeasures as well as robust attacks that can be used to evaluate the proposed countermeasures. Beyond robustness, adversarial reprogramming fills a gap in the literature on 'transfer learning by fine-tuning' by innovating socially beneficial applications [21], [59]. Model programming is a cross-domain resource-efficient ML that enables repurposing an established pre-trained model from a source domain to perform tasks in a target domain without fine-tuning (i.e., resource-efficient), For example, reprogram pre-trained English language models for protein sequence infilling [60]. As illustrated in Fig 7, model reprogramming enables cross-domain machine learning by adding two layers to a pre-trained model: 1) An input transformation layer, which is optimized to generate universal adversarial perturbation to be included in each input of a target task. This layer enables reprogramming the pre-trained model to solve the target task. In other words, it transforms an input of the target task into another representation that fits the input dimension of the pre-trained model. 2) An output mapping layer, which maps the labels of the source task to the labels of the target task.

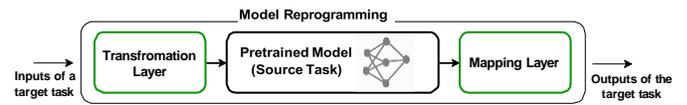

Fig. 7: Model reprogramming enables cross-domain ML by adding an input transformation as well as output mapping layers to a pre-trained model.

Model reprogramming can be particularly useful for addressing the data scarcity challenge in applications where data is limited and the acquisition and annotation of new data is often costly and time-consuming, such as healthcare.

Model reprogramming is in contrast to the typical domain adaptation problem in which the task is the same in the source and target domains. Also, model reprogramming works better than typical transfer learning (i.e., fine-tuning) because the required training data is much smaller. Most importantly, unlike fine-tuning, model reprogramming can map a source task to a different target task. This is a promising use case for leveraging adversarial ML that allows cross-domain training of ML models with small data. For instance, Tsai et al. [61] proposed to use model reprogramming for performing transfer learning while considering black-box settings, where they do not have any knowledge about the pre-trained model being used for transfer learning. Zeroth order optimization and multi-label mapping techniques are the main components of their proposed method that exploits the input-response pair of the underlying model to reprogram it in black-box settings. They demonstrated the efficacy of their proposed method by performing transfer learning for three medical imaging tasks; namely, diabetic retinopathy detection, skin cancer detection, and autism spectrum disorder detection. Similarly, model reprogramming of acoustics models for different time series prediction tasks has been performed in [62]. Specifically, the authors empirically demonstrated that model reprogramming provided a state-of-the-art performance on 19 out of 30 datasets in a famous time series prediction benchmark (i.e., UCR Archive [63]).

Furthermore, Melnyk et al. [60] leverage model reprogramming to address the lack of diversity in the generated antibody sequences. In particular, they introduce the ReprogBert framework, which is a pretrained English language model that is reprogrammed for protein sequence infilling. This framework shines as an efficient cross-domain technique that generates precise and diverse protein sequences.

*2) Scientific Discovery:* One of the fascinating and positive applications of AdvML lies in its capacity to drive scientific discoveries that could not be unrevealed otherwise. In this section, we will discuss how AdvML can be leveraged to foster scientific discoveries. In AdvML, it is very common to query the trained network to extract useful information and construct shadow models. For instance, a number of such attacks have already been demonstrated in the literature, e.g., membership inference attacks, model extraction attacks, and model inversion attacks. This notion of querying a model can be leveraged for multiple purposes. For example, Hoffman et al. [64] proposed to exploit latent embeddings of trained autoencoder via query-based strategy for molecule optimization. The proposed approach performs efficient queries to enhance the desired properties of input molecules, which are also supervised by different evaluation metrics and molecular property predictions. Furthermore, the authors empirically demonstrated that the proposed framework outperformed existing methods in the optimization of small organic molecules for solubility and drug-likeness under similarity constraints. In addition, they also showed that the proposed approach can improve two challenging tasks: (1) optimization of SARS-CoV-2 inhibitors to higher binding affinity; and (2) known antimicrobial peptides improvement towards lower toxicity. Also, the experimental results demonstrated high consistency in terms of external validations.

*B. Transparency*

DNN-based ML models are known for their limitations in explaining and reasoning the predicted decisions. This lack of transparency in DNNs requires imperative research contributions to overcome this limitation [65]–[69]. AdvML4G addresses such limitations by proposing counterfactual explanations for *Explainable Artificial Intelligence (XAI)*.

AdvML4G contributes to addressing such limitations by proposing counterfactual explanations for XAI. Counterfactual explanation (CFE) methods utilize adversarial perturbations to explain the predictions of a DL model by adversarially perturbing the original inputs to produce a different output from the model [70], thereby utilizing these adversarial perturbations to act as an ally (trying to explain the DNN decision) instead of as an adversary (trying to fool the DNN output). For example, to explain the classification of an image by a black-box model, CFE techniques try to find out what should be minimally changed in pixels of an input image to produce a significant change in the output [71]. CFE techniques have recently gained popularity due to their practicality—CFE techniques can be extended to the DL models of several architectures with arbitrarily high complexity, and do not strictly rely on the white-box access to the underlying model and human-friendly explanations, i.e., the explanations produced by CFE techniques are understandable and meaningful to the end users.

Dhurandhar et al. [72] proposed a contrastive explanation method (CEM) that, given an input, adversarially identifies the *pertinent positives*, comprising input features necessary for the final classification, and *pertinent negatives*, comprising input features that distinguish the input from the second most probable class. Similarly, Ustun et al. [73] used counterfactual explanations to generate a list of actionable changes, termed counter-factual recourse, for the benefit of the input data subjects, and present tools to evaluate the feasibility and difficulty of the recourse. Pawelczyk et al. [74] leverage a variational auto-encoder (VAE) to produce faithful counterfactual explanations from actionable training data recourses, extendable to the tabular data. Recently, Pawelczyk et al. [75] formalize the similarities between the counterfactual explanation methods and the adversarial examples generation mechanisms.

*C. Privacy*

Human privacy preservation is a fundamental right that is upheld and regulated by global governing bodies. As the deployment of AI applications that gather and analyze personal data increases, safeguarding human privacy has emerged as a critical concern. Consequently, there is a growing demand for effective countermeasures. AdvML4G contributes to addressing this need by introducing innovative applications to secure and protect human privacy. One positive application of AdvML is its implementation in *Data Cloaking* to protect privacy attributes in sensitive data. For instance, Shan et al. [76] et al. did preliminary work in this space, where they introduced small (imperceptible) adversarial perturbations into the facial images for cloaking them. Specifically, they evaluated their proposed image cloaking approach on three public cloud-based services that include Amazon Rekognition, Microsoft Azure

TABLE IV: Summary of Various AdvML4G Applications.

| Social Good Outcome | Ref | Objective | Application | Description | Evaluation | Limitation(s) |
|---|---|---|---|---|---|---|
| Innovation | [21] [59] | Cross-Domain Resource-Efficient ML | Model Reprogramming | Leveraged AdvML to reprogram a pretrained model from a source task to a different target task using a small dataset. | CIFAR-10 with pretrained ResNet18 | They can be misused by real adversaries and used for theft of computing resources or for inventing socially problematic applications. |
| | [61] | | | Used zeroth order optimization and multi-label mapping techniques to perform reprogramming in black-box setting. | Diabetic retinopathy, skin cancer, and autism spectrum disorder detection. | N/A |
| | [64] | Scientific Discovery | Molecule Optimization | Exploited querying a trained autoencoder for molecule optimization problem. | Used on a chemical dataset for toxicity prediction. | Better generative models can be explored. |
| Privacy | [76] | Enhancing Privacy | Face Recognition | Used well-known PGD attack to generate imperceptible perturbations for face image cloaking. | Amazon Rekognition, Face++, and Microsoft Azure Face. | Becomes less effective when an adversary realizes a strong targeted attack. |
| | [77] | | | Used black-box AdvML attacks to generate such facial images that can reduce chance of person re-identification. | Amazon Rekognition, and Microsoft Azure Face. | Less effective in case of strong and customized exploration of robust systems. |
| Fairness | [78] | Bias Mitigation | Domain Adaptation | Leveraged domain adaptation and adversarial training to increase model fairness. | COMPAS, Adult, Toxicity, and CelebA | Only binary classification problem is considered. |
| | [79] | | Face Recognition | Used adversarial glasses attack [80] to enhanced privacy and fairness of gender recognition model. | CelebA and the German Credit dataset | Evaluation is done only for binary classification. |
| | [81] | | Image Classification | Used PGD-based adversarial examples and boundary-based analysis for increasing fairness. | SVHN, FashionMNIST, CIFAR-10, and CIFAR-100 | Only accuracy is used for the evaluation. |
| | [82] | | Fairness Assessments | Utilized adversarial training on the client side of FL to enhance fairness. | COMPAS and Adults | focused only on individual fairness rather than other aspects of fairness. |
| Reliability and Robustness | [83] | Model Watermarking & Fingerprinting | Image Classification | Considered robustness and transferability to generate characteristics examples for fingerprinting DL models. | Base (pretrained) models, pruned models, and combination of these two. | Observed that when $k$ increases from $20-30$, uniqueness score is degraded, which affects pruned models' performance. |
| | [84] | | | Proposed GradSigns that works by embedding signature into the gradients of cost function with respect to model's input. | CIFAR-10, SVHN, and YTF | Observed a slight but negligible impact on the performance protected model |
| | [85] | Improved Generalization | Multi-tasking | Used a single adversarially *robust* classifier to solve several challenging computer vision tasks. | Imagenet, CIFAR-10 | Results were not on par with the state-of-the-art methods. |
| | [86] | | Transfer Learning | Showed that adversarially trained ML models have improved transferability to downstream tasks. | CIFAR-10, CIFAR-100, Imagenet | The best hyperparameter choice significantly depends on the dataset. |
| | [87] | | Data Augmentation | Proposed unsupervised adversarial examples (UAEs) to be used as data augmentation for the unsupervised ML tasks. | MNIST, CIFAR-10 | May fail when the training loss after augmentation is significantly higher than the original loss. |
| | [88] | Robustness Enhancement | Human Identification | Proposed DeepCAPTCHA using adversarial noise to imperceptibly perturb the standard CAPTCHAs | MatConvNet (MNIST), CNN-F (ILSVRC-2012) | Effectiveness is reduced against adversarially trained models. |
| | [89] | | Human Identification | Used visible adversarial patches to fool ML models while keeping most of the image pixels unperturbed. | VGG16, VGG19, ResNet50, InceptionV3, Xception, MobileNet | Method has not been evaluated against adversarially trained models. |
| | [90] | | Human Identification | Exploited adversarial perturbations to generate pseudo adversarial background and foreground for text CAPTCHAs that are more human-friendly than the previous methods. | KNN, SVM, LeNet, AlexNet, VGGs, GoogleNet, ResNets, DenseNets | Effectiveness is reduced against adversarially trained models. |
| | [91] | | NLP | Introduced perturbations into words vectors to generate training data for the detection of lexical semantic change. | Evaluated using three datasets of British and American English language. | Hyperparameters such as vocabulary size and no. of tokens may impact the performance. |
| Transparency | [70] | Explainable AI | Behavior Explanation | Explained the ML model output by computing adversarially perturbed inputs. | LSAT dataset, Pima Diabetes | Do not evaluate the faithfulness and feasibility of the recourses (explanations). |
| | [72] | | | Identified positive and negative features to explain the black-box model output using adversarial perturbations. | MNIST, Procurement Fraud, Brain Functional Imaging | Does not provide an actionable recourse for the agent. |
| | [73] | | Future Policy Suggestion | Used adversarial perturbations to generate a list of actionable changes for the benefit of the input data agents. | Credit dataset, Giveme Credit dataset | Only applicable to linear classification. |
| | [74] | | Tabular Data Classification | Leveraged a VAE to adversarially produce faithful (implementable) recourses (explanations) extendable to the tabular data. | Giveme Credit dataset, HELOC dataset | The most faithful recourse might not be the easiest one. |

Face, and Face++ (which are developed for facial recognition). Along similar lines, Li et al. [92] used text-based adversarial perturbations to privatize sensitive personal attributes—age, gender, and location—from the natural language processing (NLP) inference attacks.

Similarly, Cherepanova et al. [77] demonstrated that AdvML can be used to develop an adversarial filter that can hide sensitive information from the input data. Specifically, they evaluated the effectiveness of the proposed filter using Microsoft Azure Face Recognition API and Amazon Rekognition. Their proposed approach works by generating an adversarially perturbed image that lies far away from the original image in the feature space, while simultaneously minimizing a perceptual similarity loss between the perturbed and original images. The benefits of this approach are two-fold: (1) distance maximization between the perturbed and original images in the feature space prevents matching individual's other images; (2) perceptual similarity loss minimization ensures the quality of the generated image is not degraded. There have been many other proposed works that defend ML-powered systems against privacy violations [93].

In contrast to the abovementioned works (i.e., Fawkes and LowKey), Cilloni et al. [94] proposed the Ulixes approach, which employs adversarial examples to generate imperceptible noise masks (cloaks) that are computationally efficient and thus can be used by end-users to preserve their privacy. In particular, users cloak their images before uploading them to online services, and thus, the trained face recognition systems fail to identify the real identity associated with those images. Ulixes utilizes the transferability of adversarial attacks to generate cloaks that are effective in various face recognition systems. It fits in scenarios where an end-user cannot poison datasets that might include some of their images as in the case of Fawkes and Lowkeys works. Ulixes differs from these works on the threat model and the cost. They require poisoning the dataset, which is computationally inefficient, while Ulixes requires cloaking some of the images, which is computationally efficient.

Due to the ubiquity of the diffusion models that enable users to generate new images given a prompt and/or target image, the risk of maliciously utilizing these models has become a pressing privacy concern. The study by Salman et al. [10] addresses this concern by employing adversarial perturbations. They introduced two variants of adversarial attacks, namely encoder attacks and diffusion attacks, which can be applied before users upload their images online. These attacks serve to immunize the images against malicious usage such that malicious users cannot utilize these images as inputs to diffusion models to generate offensive and realistic image variations. Given an original image to be immunized as well as a target image, the encoder attack aims to map the representation of the original image to the representation of the given target image. In contrast, the proposed diffusion attack aims to break the diffusion process by manipulating the whole process to generate an image that resembles the target image.

Private information on social media profiles can be leaked or collected without explicit permission. AdvML4G addresses this issue through an adversarial attack that generates graphs and adversarial features [20]. In particular, features on the edges and nodes are perturbed such that Graph Neural Networks cannot recognize private information. Deepfake attacks have become a real threat to individual privacy. Few works utilize AdvML4G to proactively defend against Deepfake threats before data manipulation. For example, Yang et al. [95] proposed transformation-aware adversarial faces that hinder crafting high-quality fake outputs. Another example is the work of Want et al. [96], which defends against deepfake by generating perceptual-aware perturbations that are resilient to various input transformations. Similarly, He et al. [97] protect against deepfake by utilizing an encoder that transforms real faces into latent space and searches the embeddings of the adversarial face in their neighbor with the gradient descent method.

Furthermore, AdvML4G can be used for emotional privacy preservation. There are situations where unauthorized organizations violate the privacy rights of individuals and utilize Emotional Recognition Systems (ERS) to detect emotions. AdvML4G can be used to preserve emotional privacy in such scenarios. Shawqi et al. [98] proposed an approach called Chaining Adversarial Attacks (CAA) that robustifies adversarial attacks to be utilized as allies to evade unauthorized ERS. In particular, CAA aims to enhance the robustness of adversarial examples by passing the targeted emotion (facial expression image) through a pipeline comprising a sequence of stages. In this pipeline, the output of a previous stage (i.e., an adversarially attacked emotion) is used as an input to a subsequent stage, subjecting the targeted emotion to another round of adversarial attack. The result of this chaining of attacks is a robustly adversarially attacked emotion. This can be employed by victim users as an ally attack to safeguard their emotional privacy from unauthorized emotion detection systems.

### D. Fairness

It is widely recognized that the biases in training data get transferred to the associated trained model, which results in the development of biased models lacking fairness in their predictions. Among other methods for enhancing the fairness of DNN-based models, AdvML can potentially contribute to mitigating bias and ensuring fairness in DNN-based model predictions.

For instance, Liang et al. [78] proposed utilizing domain adaptation and adversarial training to improvise a fair classification model. Specifically, they presented FAIRDA framework, which transfers knowledge from a source domain to adapt sensitive attributes (including those unknown in the source domain, such as race and gender) and learn a fair model in the target domain. This frameworkhas two main components: (1) an adversarial domain adaptation module responsible for estimating sensitive attributes for the target domain, and (2) an adversarial debiasing module aimed at training a fair model for the target domain.

Similarly, the Federated Learning (FL) framework, as a variant of ML, suffers from bias issues. AdvML contributes to addressing this concern by proposing a few works. For instance, Li et al. [82] utilize adversarial training on the client side to enhance fairness in federated learning. Specifically, they enable individual fairness (i.e., a model that treats similar inputs similarly) by defining a similarity metric for individual fairness, generating examples that violate the established definition of individual fairness, and then utilizing those examples to perform adversarial training on all FL clients. This process results in a model that satisfies individual fairness.

AdvML4G can also be used to enhance fairness and privacy at the same time. The work of Belavadi et al. [79] presents a method that protects the privacy of individuals and offers them a fair chance to play a legitimate game with the ML system by motivating the enhancement of the user's profile with positive updates so that they can overcome potential existing bias.

DNN-based models learn more for the majority class. Thus, they become biased towards this class. AdvML4G can contribute to addressing such an issue. For example, Zhang et al. [81] used the PGD attack algorithm to generate adversarial examples lying close to the decision boundary in order to guide the training of the DNN so as to notably alleviate the negative effects of class imbalance in the training data. More specifically, the authors showed that guided adversarial training significantly improves the accuracy of the DNN over those classes that have a relatively small representation in the training set.

*E. Reliability and Robustness*

AdvML4G applications contribute to building reliable and robust ML-powered systems, which are relevant to human safety. It is worth noting that, although robustness is a main aspect of AdvML (i.e., adversarial robustness aspect), it was introduced in a purely technical context that was meant to fix identified performance issues in DNNs. However, in AdvML4G, we introduce robustness in a social context, where robustness is required for societal well-being. Insecure applications are unsafe as adversaries can potentially exploit them to launch negative impacts on society. Thus, for the safety and security of users, the utilized systems should be robust and reliable.

*1) Model Watermarking and Fingerprinting:* Watermarking and fingerprinting are crucial for developing approaches that protect model vendors against intellectual property infringements. Also, with the increasing popularity of pre-trained models, it is important to ensure ownership of these models [83]. However, existing watermarking techniques are vulnerable to watermark removal attacks. AdvML can be used for improvising novel watermarking frameworks for developing DNNs that are robust against watermark removal methods. In this regard, Wang et al. [83] proposed the use of characteristic examples to fingerprint DL models. Specifically, they considered the joint problem of robustness and transferability to generate realistic fingerprints and proposed three types of characteristic examples that include (1) C-examples; (2) RC-examples; and (3) LTRC-examples. These examples were used to derive fingerprints from the base (original) model. In addition, to address the tradeoff between transferability and robustness, they proposed the Uniqueness Score metric that quantifies the difference between transferability and robustness and also helps with the identification of false alarms.

In a similar study [84], the authors proposed a novel model watermarking technique named GradSigns that works by embedding signatures into the gradients of the cost function with respect to the input during model training. The intuition of GradSigns is based on the fact that DL models can find more than one solution to the non-convex problems they solve. Therefore, GradSigns works by finding such a solution (set of model parameters) that satisfies two properties, i.e., finding the decision boundary for the given task with comparable performance and embedding the owner's signature (i.e., watermark information) into the model parameters. The proposed method was evaluated on image classification tasks using various datasets that include CIFAR-10, SVHN, and YTF. Moreover, GradSigns was found robust against existing watermark removal attacks, highlighting the efficacy of the proposed approach. However, the authors observed a slight (negligible) impact on the performance-protected model. Also, GradSigns can be used for the remote verification of watermarks by DL models' vendors using prediction APIs. Sablayrolles et al. [99] proposed the use of radioactive data to identify whether an input was used for training or not. The proposed method works by making imperceptible changes to radioactive data such that the model being trained on this data provides an identifiable signature. The proposed approach is capable of working in both white-box and black-box settings. Although this approach significantly differs from watermarking or fingerprinting DL models, this approach demonstrates the use of AdvML to watermark datasets in a classical sense [100].

*2) Model Generalization Improvement:* Adversarial training is performed by augmenting the training dataset with adversarially perturbed inputs. Many recent works have shown that adversarial training improves the generalization and robustness of ML models. Santurkar et al. [85] used a single *robust* classifier, adversarially trained on some dataset, to solve several challenging computer vision tasks in the manifold of the dataset that the robust classifier is trained on. These tasks include image generation, image inpainting, image-to-image translation, super-resolution, and interactive image manipulation. Conventional ML methods require different ML models of varying architectures trained on different loss functions to solve each of the aforementioned tasks. Furthermore, they showed that all of these tasks can be reliably and single-handedly performed by a single robust classifier by simply perturbing the input image to maximize the targeted class probability. Salman et al. [86] empirically show that training ML models on adversarial examples improve their transferability to downstream tasks, mainly because of the improved feature representations enabled by the adversarial training. Hsu et al. [87] proposed a method to generate unsupervised adversarial examples (UAEs) and use the proposed method as a data augmentation tool for several unsupervised ML tasks. Lee et al. [101] adversarially manipulated the training data images to increase the confidence score of a classifier. The input attribution map generated by the gradients of the classifier was then used to improve semi-supervised semantic segmentation models.

*3) Robustness Enhancement:* Completely Automated Public Turing Test to Tell Computers and Humans Apart (CAPTCHA) are widely used by several applications to distinguish between human and robot users. Osadchy et al. [88] note that CAPTCHAs and adversarial examples have a shared objective; they are challenging to be recognized by the automated systems (such as ML models), while significantly easy for humans to solve. The authors, therefore, propose DeepCAPTCHA that uses *immutable* adversarial noise, robust to filtering or other preprocessing mechanisms, to imperceptibly perturb the standard CAPTCHAs. On similar grounds, Shi et al. [102] propose four text-based and four image-based adversarial CAPTCHA generation methods. Noting that adversarial perturbations may noticeably impact the semantics of the image or text CAPTCHAs even for human observers, Hitaj et al. [89] use visible adversarial patches to fool ML models while keeping most of the image pixels unperturbed so as not to affect human-friendliness of the CAPTCHAs. Shao et al. [90] exploit adversarial perturbations to generate pseudo-adversarial background and foreground for text CAPTCHAs that are more human-friendly than the previously proposed adversarial CAPTCHAs. Recently, Zhang et al. [103] robus-tify adversarial CAPTCHAs by considering issues such as sequential recognition, indifferentiable image preprocessing, stochastic image transformation, and black-box cracking to produce adversarial examples.

Inspired by well-designed perturbations (adversarial examples) in the image classification task, Hammed et al. [104] suggested robustifying wireless communication channels against adversaries (intruders) by crafting well-designed perturbations that are strong enough to be reliably decoded by legitimate users while fooling adversaries that may intercept the transmitted signals. Specifically, small modifications are introduced to the modulation scheme (in-phase/quadrature symbols) at the transmitter to fool the classifier employed by potential adversaries. Consequently, adversaries are unable to identify the scheme, which makes it challenging for them to decode the underlying information. This approach can be crucial in scenarios where strong encryption is unfeasible, such as in the limited computational resources of IoT devices.

Furthermore, the work of Salman et al. [105] utilizes adversarial examples to robustify the object (e.g., image) itself, not the model. They utilize the concept of adversarial examples to design un-adversarially perturbed objects that are optimized to be confidently classified. Moreover, Cresci et al. [106] demonstrate that adversarial training can contribute to robustifying the detection of fake news and social bots by improving the robustness of fake news and social bot detectors.

Budgeted Human-in-the-Loop (HitL) can be effectively integrated during both the training and inference stages to enhance the robustness and reliability of AdvML-based systems. Training ML models using budgeted HitL can significantly contribute to the development of socially beneficial applications that are both robust and reliable. The incorporation of HitL during the training phase enables the creation of resilient online systems against various distribution shifts, including adversarial and natural shifts. A work by Al-Malik et al. [107] exemplifies this by adopting a systematic active fine-tuning approach, which emerges as an efficient and cost-effective method for producing online ML systems capable of handling distribution shifts effectively. Similarly, the integration of HitL during deployment holds the potential to enhance socially beneficial (robust and reliable) AdvML-based applications. Al-Malik et al. [108] propose a strategy that augments State-of-the-Art (STOA) adversarial defense approaches with an online selection and relabeling algorithm (OSRA). This innovative approach aims to improve the robustness of deep learning-based systems against potential adversarial attacks.

AdvML can contribute to detecting language variations. It is very common to see variations in the use of language across different knowledge domains and demographics, which often happens in monolingual societies as well. This language change is also known as lexical semantic change (LSC), which is concerned with the quantification and characterization of variations in language with respect to its semantic meaning. In natural language processing, LSC is considered to be quite challenging due to the unavailability of representative (application-specific) databases. Therefore, the majority of literature focused on LSC uses unsupervised learning-based approaches to detect language variations (semantic shift) in embedding space. To address these issues leveraging AdML4G, Gruppi et al. [91] proposed a self-supervised learning-based approach that generates training samples to model LSC. Specifically, they proposed to introduce perturbations in word embeddings in the input to generate training data and demonstrated that the proposed approach can be integrated with any alignment method to detect semantic shifts. They evaluated the proposed approach against semantic shifts in British and American English languages using British National Corpus (BNC) XML Edition [109] and Corpus of Contemporary American English (COCA) [110] datasets, respectively.

*F. Discussions and Insights*

ML application developers have enough motivation to build *anti-social* applications because these applications are more profitable. Therefore, it is crucial to confront these *anti-social* applications by establishing effective regulation and motivating ML developers to build ML4G applications - that is, encouraging them to build pro-social applications and mitigate socially harmful ones. Unfortunately, pro-social application relevant regulations are currently immature and have many limitations [111]. Thus, the role of ML4G developers is essential as an effective solution that complements the limitations of pro-social application regulations. To this end, mitigating the negative impact of anti-social applications can be performed by leveraging prompt yet effective ML4G enablers, such as AdvML4G. AdvML4G shines as an auxiliary ML tool that enables ML4G. In addition, it effectively complements the limitations in pro-social application regulations whenever conventional ML tools do not work. For example, affected individuals may attempt to defend themselves by reporting the incidents to authorities to initiate the appropriate actions. However, this defensive approach has various limitations. First, thoughtful regulations handling technology misuse are immature and not well-established everywhere. Moreover, even if they are established, proving AI misuse incidents is challenging as they are often performed by experts in incident trait clearance. In addition, this approach requires a lot of resources, such as workforce, software, and other tools. Second, the authority's actions against technology misuse usually take time due to the associated validation and tracking process. Defending technological incidents with such conventional and manual actions is ineffective. AI technology misuse requires prompt and effective actions to bring them into line with social responsibility. More specifically, authorities and affected individuals can reactively utilize AdvML4G to defend against the anti-social impacts of misused technology. In addition, ML4G developers can proactively augment their *pro-social* tools with AdvML4G as an auxiliary tool to innovate more socially beneficial applications effectively.

In addition to the discussion mentioned above, below are a few insights from exploring AdvML4G applications:

- AdvML4G applications typically aim to prevent/mitigate the negative impact of anti-social applications or facilitate the development of pro-social ones.
- Adversarial robustness, as an unsolved AdvML problem, can be utilized for social good to enhance human safety.

For example, it can be utilized as a trust enabler for differentiating humans and machines (i.e., human authentication) as bots and machines are easily deceived by evasion attacks, while humans are not.
- The larger proportion of published adversarial attacks compared to the published defenses [22] indicates that pro-social application developers are armed with enough tools to increase innovation.
- The most impactful adversarial attacks for innovating pro-social applications are adversarial reprogramming attacks. In contrast, evasion, as well as poisoning attacks, are the most common and impactful attacks utilized for mitigating harmful impacts.

## V. CHALLENGES OF ADVML4G

### A. Quantifying Intent of AdvML Attacks

AdvML attacks can be potent tools potentially having both socially beneficial and harmful outcomes. Therefore, understanding and quantifying the intent behind these attacks poses a multifaceted challenge. Since many attacks are application agnostic and can be applied for various purposes. For instance, an attack designed to achieve a positive outcome can be utilized to get malicious outcomes, e.g., an attack for enhancing privacy may also be employed for malicious surveillance. This inherent duality of AdvML poses a significant challenge in determining whether a specific attack is socially good or harmful. Furthermore, it is worth noting that it is significantly challenging to objectively tell whether the purpose of an algorithm classifies as good or not, due to the subjectivity of the observer and the ever-evolving definition of the good—what today classifies as a not-good application may become good in the future and vice versa.

### B. Limitations of AdvML4G Applications

Although AdvML algorithms can achieve several desired objectives in numerous applications, these algorithms suffer from some major limitations in most of these applications. For example, Shawn et al. [76] proposed Fawkes that uses adversarially poisoned facial images to protect users' privacy from online facial recognition models. However, Radiya et al. [112] demonstrated that Fawkes fails under a practical setting because an attacker who appears in the future can use these poisoned images to train a better model or use new state-of-the-art techniques not available at the time of data poisoning to circumvent the protection. Similarly, counterfactual explanations [70], which we believe are one of the most impactful applications of AdvML4G, may not either truly explain the DNN behavior or produce practically feasible recourses for the stakeholders to make modifications in the input in order to change the DNN output [74]. Overall, many novel applications of AdvML are still in their infancy, and, therefore, have not gained much popularity yet. We are hopeful that as future developments address these limitations, AdvML4G will widely attract the attention of AdvML researchers.

### C. Reverse Engineering and Defense Countermeasures

Despite the fact that: 1) AdvML4G applications are not assumed to be deployed in adversarial settings (i.e., security-critical scenarios), where strong defenses are considered essential; and 2) adversarial robustness remains an unsolved problem, meaning that highly effective defenses have not been proposed yet, there is still a possibility that, in white-box threat models where the tools used for developing the application are known, AdvML defenses could be utilized as attacks (defense-as-attacks) to reduce the effectiveness of the deployed AdvML4G applications. For example, applications exploiting AdvML4G are exposed to the reverse engineering threat and mitigating countermeasures. In other words, state-of-the-art countermeasures against adversarial attacks, such as adversarial training [2] and certified smoothing [3], can hinder the efficacy of AdvML4G. For example, adversarial reprogramming attack-as-ally can be mitigated by some defenses such as stateful detection [113]. The effect of these defenses is limited, though.

### D. Lack of Researcher-Practitioner and Policy-Makers Collaboration

Collaboration between researchers and practitioners in the AdvML community is vital for practical research outcomes. However, currently, this collaboration is inadequate. For example, there are no real-world open-source systems for researchers to validate their proposals. Evaluating the proposed pro-social applications in real-world settings is the proper strategy to ensure quality and effective outcomes. Similarly, researchers and practitioners lack collaboration with policy-makers.

### E. Regularizing Adversaries as Allies

Regularizing AdvML might be a challenge because we only distinguish between an adversary and an adversary-as-ally based on the intent. Determining whether a given application is *good* or not is a major challenge due to the subjectivity and the lack of suitable benchmarks to evaluate and quantify the *good*ness of an application. Yet another significant challenge is formalizing the identification of adversary-as-ally, which can be critical in future policy making. In other words, the challenge here lies in how regulators can recognize the adversary-as-ally to grant him permission to develop AdvML4G. These challenges may call for utilizing adversarial attacks-as-ally as a contingency solution when the normal approaches are not working. Otherwise, since such pro-social applications are on the edge, they can easily get out of control and be used as a socially offensive weapon. The most restrictive challenges in developing AdvML4G include regularizing adversaries as allies and addressing the absence of well-established benchmarks and performance metrics that facilitate the evaluation of developed AdvML4G applications.

## VI. FUTURE RESEARCH DIRECTIONS
### A. Developing Proxy Metrics for Quantifying Intent

As discussed earlier, it is quite challenging to quantify the intent of AdvML attacks due to the intricate interplay of positive and potentially harmful outcomes of these attacks. Therefore, the development of proxy metrics is crucial which can provide a quantitative lens through which the intent of AdvML attacks can be assessed. To be specific, various factors can be considered for developing proxy metrics including adversarial objectives and the potential societal impact of the developed attack(s). Moreover, these metrics can be augmented with ethical principles and guidelines that could ensure a more objective and consistent means of evaluating the intent of AdvML attacks. However, the development of such quantitative metrics is quite challenging and the subjectivity and dynamic nature of ethics remain a persistent concern. Therefore, it demands interdisciplinary research efforts to develop effective and robust proxy metrics to quantify the intent of AdvML attacks.

*B. New Kinds of Adversarial Attacks as Allies*

Most AdvML4G application developers pay more attention to utilizing evasion [1], poisoning [13], and adversarial reprogramming [14]. We believe other adversarial attacks can be utilized as allies and contribute to AdvML4G. We provide two examples next.
- *Membership extraction* [46] *attack-as-ally* can be used in AdvML4G for building privacy-violation inspection applications. In the absence of a responsible authority that protects the collection and processing of private human information, the societal role of the adversary-as-ally to identify and report information privacy violations is imperative. This role can also complement the authorities' role when their usual solutions don't work.
- *Model stealing* [45] *attack-as-ally* can be utilized in AdvML4G in scenarios where anti-social applications cannot be stopped (e.g., due to their deployment location). Such an attack-as-ally can perform model stealing to identify their internals as a preliminary step to launch another type of adversarial attack that ruins or stops those offensive applications.

*C. Establishing AdvML4G Benchmarks and Standards*

To ensure the healthy growth of the emerging field of AdvML4G, it is crucial for researchers and practitioners to collaborate in developing benchmarks, datasets, and best-practice guidelines specifically tailored for building, training, and evaluating AdvML4G applications. This collaborative effort will provide a solid foundation and standardized framework, enabling the advancement and widespread adoption of AdvML4G solutions.

*D. Pro-Social applications enabled through multiple AdvML attacks*

Most AdvML attacks are crafted utilizing one type of attack (i.e., one distance metric), which is the de facto standard in the AdvML (adversarial robustness aspect) ( [114]). However, AdvML4G should go beyond and explore employing multiple AdvML attacks while innovating socially good applications or mitigating socially harmful ones. We strongly encourage researchers and practitioners in the AdvML4G community to adopt the utilization of multiple attacks to innovate socially beneficial applications.

*E. AdvML4G and smart city applications*

Smart city applications should be augmented to enable social good outcome. Conventional smart city applications promote the well-being of individuals and communities but not necessarily considering human values. We encourage collaboration between AdvML4G and smart city communities. This collaboration can utilize AdvML attacks as allies to 1) innovate new smart city applications that conventional ML development tools cannot produce, and 2) mitigate smart city applications that violate human values.

*F. AdvML4G and Generative AI Applications*

Due to their advanced capabilities, applications of generative AI, such as LLMs-based systems and diffusion models, have a higher potential of being misaligned with human values As a result, it is crucial for researchers to engage in initiatives aimed at reducing or mitigating this risk of misalignment. Given its distinct ability to innovate socially beneficial applications and mitigate socially harmful ones that would otherwise remain unaddressed, AdvML4G holds substantial potential to contribute towards aligning these systems with human values. Collaborative work between researchers in generative AI and AdvML4G communities can pave the way for augmenting these systems with AdvML4G-based capabilities that enable a more ethical and value-aligned future in AI systems. In addition, AdvML4G can maximize the potential of generative AI Models. For example, AdvML4G tools, such as model reprogramming, can enable generative AI models, such as diffusion models and LLMs, to maximize the utilization of their capabilities by empowering data-limited domains to develop a wide range of useful downstream applications.

*G. Developing AdvML4G Applications Using Unsupervised Learning*

Current AdvML4G tools, such as model reprogramming, rely on supervised learning. In other words, model reprogramming employs a small labeled dataset to reprogram (adapt) a pre-trained model (source task) to a new target task. It is crucial to investigate the potential impact of unsupervised learning on the effectiveness of the reprogramming process. With advancements in unsupervised learning tools, there is a significant potential for improvement.

*H. AdvML4G and Efficient Distributed Training for Downstream Applications*

Certain implementations of AdvML4G tools, such as model reprogramming, can reduce communication time and costs, leading to increased productivity and operational cost savings. An example of such an implementation is deploying a model reprogramming-based system as a distributed ML system, where an input transformation layer and output mapping layer are placed on edge devices, while the frozen pre-trained model is hosted on a central server.

## VII. Roles of Governments, Industry, and Academia in Promoting Pro-Social ML

Collective efforts of governments, academia, and industry are necessary to push toward more pro-social and less anti-social ML applications. Governments are requested to establish the rules and policies that regulate the development of ML applications. Academia bears the role of starting new research directions that are pro-social in nature. The industry plays the role of producing systems that are pro-social by design. 8 illustrates how the collective role of governments, academia, and industry contributes to facilitating the development of more ML4G applications, and consequently, more *AdvML4G*. Further details on the recommended roles for government, academia, and industry are presented in the following subsections.

### A. Governance for Pro-Social ML

The current ML economic ecosystem is profit-oriented. It motivates entities involved in developing ML applications to pay more attention to the applications that maximize profit. This ecosystem results in inadequate innovation in areas with socially beneficial impacts but with low-profit value. In the absence of motivation towards socially beneficial applications and in the lack of established guides and policies that ensure the development of ML applications that are socially good, ML system developers tend to build profitable applications regardless of their social impact. The tragedy of the commons economic theory supports this tendency [111]. Another supporting evidence is the latest report from Human-Centered AI (HAI), which highlights that anti-social AI applications are on the rise [115]. Thus, governments should direct the development of ecosystems and close the gap by establishing policies and offering governance and strategic incentives that encourage AI developers to build pro-social applications and abstain from developing anti-social ones.

The need to establish proper governance is particularly urgent in light of the significant advances in Large Language Models (LLMs), such as ChatGPT and similar variants. The potential for AI to be utilized for producing socially problematic applications at scale is now a profound risk. To address this issue, AI business leaders and top researchers have signed an open letter requesting that AI companies competing to build ChatGPT-like or any human-competitive intelligence systems pause their development for six months [116]. This request aims to provide policymakers and regulators around the world with sufficient time to develop policies and guidelines that ensure healthy progress in the AI field while mitigating the potentially catastrophic impact of this technology on humanity. This may include restrictions on the development of socially negative applications.

Ensuring that contemporary AI systems have a positive impact on society and humanity requires not only proper regulations and governance but also the efforts of experts in industry and academia across various AI communities. These experts must work to develop responsible and human-centered ML systems that uphold fundamental human rights.

### B. Roles of Industry in Pro-Social ML

The role of industry to help in developing pro-social applications can be summarized as follows: vendors (as industry experts) should: (1) commit to producing applications that comply with UN Guiding Principles (UN Guiding Principles on Business and Human Rights) [117]; (2) comply with cross-industry standards that ensure transparency and human rights preservation; (3) blocklist or allowlist customers based on their human rights preservation record; and (4) adopt a human right-by-design production strategy, in which vendors hinder the deployment of anti-social applications. For example, they incorporate the produced systems with features that trigger alerting regulators and responsible organizations whenever an anti-social use case is detected [23].

### C. Roles of Academia in Pro-Social ML

Similar to the recommended role of industry experts, leading academic researchers in different AI communities should chart new directions that prioritize innovating socially beneficial systems, either by developing intrinsically positive systems or by offsetting the socially negative impact of other systems. To achieve this, distinguished researchers in AdvML community are repurposing adversarial ML attacks to go beyond adversarial robustness, among other motivations, and embrace a socio-technical dimension. Overall, a collaborative effort from experts across different AI domains is vital for ensuring AI systems align with societal values and contribute to human well-being. In addition, specific collaborations among these entities can support pro-social ML. For instance, governments can partner with universities to develop techniques tailored to local laws and regulations when general solutions may not apply directly.

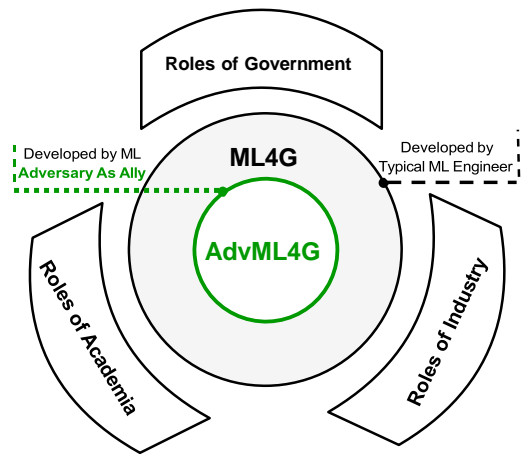

Fig. 8: The collective role of Governments, Industry, and Academia contributes to facilitating the development of more ML4G applications, and consequently, more AdvML4G applications.

## VIII. Conclusions

Deep Neural Networks (DNNs) have demonstrated vulnerability to various types of adversarial attacks, necessitating the development of AdvML as a research field to enhance the robustness of DNNs against such threats. While AdvML researchers have made significant contributions to strengthening DNNs, the proposed attacks and defenses have shown limited efficacy in real-world settings, primarily proving effective only in laboratory environments. The industry initially overlooked adversarial attacks as a genuine threat, creating a perception that hindered the adoption of adequate defenses. This limitation has sparked contemplation within the AdvML community regarding the feasibility of further investment in traditional approaches. Concurrently, AdvML holds untapped potential beyond defense mechanisms. It encompasses neutral tools that can be employed positively, utilizing adversarial attacks as allies to drive innovation and develop socially responsible applications. Consequently, AdvML4G has emerged as a new research direction that aims to harness the positive implications of adversarial attacks for the development of socially responsible applications. The existing literature lacks a comprehensive work that introduces, motivates, provides recommendations, and summarizes the applications within this promising field. This work aims to bridge that gap by offering a holistic exploration of AdvML4G and its implications.

## IX. Acknowledgment

This publication was made possible by NPRP grant # [13S-0206-200273] from the Qatar National Research Fund (a member of Qatar Foundation). The statements made herein are solely the responsibility of the authors.

## References

[1] C. Szegedy, W. Zaremba, I. Sutskever, J. Bruna, D. Erhan, I. Goodfellow, and R. Fergus, "Intriguing properties of neural networks," *arXiv preprint arXiv:1312.6199*, 2013.

[2] A. Madry, A. Makelov, L. Schmidt, D. Tsipras, and A. Vladu, "Towards deep learning models resistant to adversarial attacks," *arXiv preprint arXiv:1706.06083*, 2017.

[3] J. Cohen, E. Rosenfeld, and Z. Kolter, "Certified adversarial robustness via randomized smoothing," in *International Conference on Machine Learning*. PMLR, 2019, pp. 1310–1320.

[4] H. Ali, M. S. Khan, A. AlGhadhban, M. Alazmi, A. Alzamil, K. Alutaibi, and J. Qadir, "Con-detect: Detecting adversarially perturbed natural language inputs to deep classifiers through holistic analysis," *Computers & Security*, p. 103367, 2023.

[5] N. Carlini. Adversarial Attacks That Matter. [Online]. Available: https://nicholas.carlini.com/talks

[6] F. Tramer. Does Adversarial Machine Learning Research Matter? [Online]. Available: https://floriantramer.com/talks/

[7] P.-Y. Chen. Adversarial machine learning for good. [Online]. Available: https://sites.google.com/view/advml4good

[8] Z. Kolter. Robustness in Machine Learning: A Five-Year Retrospective. [Online]. Available: https://www.youtube.com/watch?v=jm4pfAP_hPs&embeds_referring_euri=https%3A%2F%2Fsatml.org%2F&source_ve_path=MjM4NTE&feature=emb_title

[9] J. Cowls, A. Tsamados, M. Taddeo, and L. Floridi, "A definition, benchmark and database of AI for social good initiatives," *Nature Machine Intelligence*, vol. 3, no. 2, pp. 111–115, 2021.

[10] H. Salman, A. Khaddaj, G. Leclerc, A. Ilyas, and A. Madry, "Raising the cost of malicious ai-powered image editing," *arXiv preprint arXiv:2302.06588*, 2023.

[11] K. Albert, M. Delano, B. Kulynych, and R. S. S. Kumar, "Adversarial for good? how the adversarial ml community's values impede socially beneficial uses of attacks," in *ICML 2021 Workshop on Adversarial Machine Learning*.

[12] P.-Y. Chen and S. Liu, "Holistic adversarial robustness of deep learning models," in *AAAI Conference on Artificial Intelligence*, 2023.

[13] B. Biggio, B. Nelson, and P. Laskov, "Poisoning attacks against support vector machines," *arXiv preprint arXiv:1206.6389*, 2012.

[14] G. F. Elsayed, I. Goodfellow, and J. Sohl-Dickstein, "Adversarial reprogramming of neural networks," in *International Conference on Learning Representations*, 2018.

[15] I. J. Goodfellow, J. Shlens, and C. Szegedy, "Explaining and harnessing adversarial examples," *arXiv preprint arXiv:1412.6572*, 2014.

[16] A. Nguyen, J. Yosinski, and J. Clune, "Deep neural networks are easily fooled: High confidence predictions for unrecognizable images," in *Proceedings of the IEEE Conference on Computer Vision and Pattern Recognition*, 2015, pp. 427–436.

[17] B. Biggio, I. Corona, D. Maiorca, B. Nelson, N. Šrndić, P. Laskov, G. Giacinto, and F. Roli, "Evasion attacks against machine learning at test time," in *Joint European Conference on Machine Learning and Knowledge Discovery in Databases*. Springer, 2013, pp. 387–402.

[18] A. Kurakin, I. Goodfellow, and S. Bengio, "Adversarial machine learning at scale," *arXiv preprint arXiv:1611.01236*, 2016.

[19] N. Papernot, P. McDaniel, X. Wu, S. Jha, and A. Swami, "Distillation as a defense to adversarial perturbations against deep neural networks," in *2016 IEEE Symposium on Security and Privacy (SP)*. IEEE, 2016, pp. 582–597.

[20] C. Kumar, R. Ryan, and M. Shao, "Adversary for social good: Protecting familial privacy through joint adversarial attacks," in *Proceedings of the AAAI Conference on Artificial Intelligence*, vol. 34, no. 07, 2020, pp. 11 304–11 311.

[21] A. Chen, P. Lorenz, Y. Yao, P.-Y. Chen, and S. Liu, "Visual prompting for adversarial robustness," in *ICASSP 2023-2023 IEEE International Conference on Acoustics, Speech and Signal Processing (ICASSP)*. IEEE, 2023, pp. 1–5.

[22] G. Apruzzese, H. S. Anderson, S. Dambra, D. Freeman, F. Pierazzi, and K. A. Roundy, ""real attackers don't compute gradients": Bridging the gap between adversarial ML research and practice," *arXiv preprint arXiv:2212.14315*, 2022.

[23] K. Albert, J. Penney, B. Schneier, and R. S. S. Kumar, "Politics of adversarial machine learning," *arXiv preprint arXiv:2002.05648*, 2020.

[24] B. Biggio and F. Roli, "Wild patterns: Ten years after the rise of adversarial machine learning," *Pattern Recognition*, vol. 84, pp. 317–331, 2018.

[25] A. Qayyum, M. Usama, J. Qadir, and A. Al-Fuqaha, "Securing connected & autonomous vehicles: Challenges posed by adversarial machine learning and the way forward," *IEEE Communications Surveys & Tutorials*, vol. 22, no. 2, pp. 998–1026, 2020.

[26] F. Khalid, H. Ali, M. A. Hanif, S. Rehman, R. Ahmed, and M. Shafique, "Fadec: A fast decision-based attack for adversarial machine learning," in *2020 International Joint Conference on Neural Networks (IJCNN)*. IEEE, 2020, pp. 1–8.

[27] W. Brendel, J. Rauber, and M. Bethge, "Decision-based adversarial attacks: Reliable attacks against black-box machine learning models," *arXiv preprint arXiv:1712.04248*, 2017.

[28] P.-Y. Chen, H. Zhang, Y. Sharma, J. Yi, and C.-J. Hsieh, "Zoo: Zeroth order optimization based black-box attacks to deep neural networks without training substitute models," in *Proceedings of the 10th ACM workshop on artificial intelligence and security*, 2017, pp. 15–26.

[29] E. Wong, F. Schmidt, and Z. Kolter, "Wasserstein adversarial examples via projected sinkhorn iterations," in *International Conference on Machine Learning*. PMLR, 2019, pp. 6808–6817.

[30] S. Mei and X. Zhu, "Using machine teaching to identify optimal training-set attacks on machine learners," in *Proceedings of the aaai Conference on Artificial Intelligence*, vol. 29, no. 1, 2015.

[31] P. W. Koh and P. Liang, "Understanding black-box predictions via influence functions," in *International Conference on Machine Learning*. PMLR, 2017, pp. 1885–1894.

[32] L. Muñoz-González, B. Biggio, A. Demontis, A. Paudice, V. Wongrassamee, E. C. Lupu, and F. Roli, "Towards poisoning of deep learning algorithms with back-gradient optimization," in *Proceedings of the 10th ACM workshop on artificial intelligence and security*, 2017, pp. 27–38.

[33] C. Xie, Y. Long, P.-Y. Chen, and B. Li, "Uncovering the connection between differential privacy and certified robustness of federated learning against poisoning attacks," *arXiv preprint arXiv:2209.04030*, 2022.

[34] M. Bober-Irizar, I. Shumailov, Y. Zhao, R. Mullins, and N. Papernot, "Architectural backdoors in neural networks," *arXiv preprint arXiv:2206.07840*, 2022.

[35] T. Gu, K. Liu, B. Dolan-Gavitt, and S. Garg, "Badnets: Evaluating backdooring attacks on deep neural networks," *IEEE Access*, vol. 7, pp. 47 230–47 244, 2019.

[36] M. Sun and Z. Kolter, "Single image backdoor inversion via robust smoothed classifiers," *arXiv preprint arXiv:2303.00215*, 2023.

[37] X. Chen, C. Liu, B. Li, K. Lu, and D. Song, "Targeted backdoor attacks on deep learning systems using data poisoning," *arXiv preprint arXiv:1712.05526*, 2017.

[38] T. A. Nguyen and A. Tran, "Input-aware dynamic backdoor attack," *Advances in Neural Information Processing Systems*, vol. 33, pp. 3454–3464, 2020.

[39] H.-Y. Lin and B. Biggio, "Adversarial machine learning: Attacks from laboratories to the real world," *Computer*, vol. 54, no. 5, pp. 56–60, 2021.

[40] N. Carlini and D. Wagner, "Towards evaluating the robustness of neural networks," in *IEEE Symposium on Security and Privacy (SP)*, 2017, pp. 39–57.

[41] N. Kumar, S. Vimal, K. Kayathwal, and G. Dhama, "Evolutionary adversarial attacks on payment systems," in *2021 20th IEEE International Conference on Machine Learning and Applications (ICMLA)*. IEEE, 2021, pp. 813–818.

[42] A. Qayyum, J. Qadir, M. Bilal, and A. Al-Fuqaha, "Secure and robust machine learning for healthcare: A survey," *IEEE Reviews in Biomedical Engineering*, vol. 14, pp. 156–180, 2020.

[43] S.-M. Moosavi-Dezfooli, A. Fawzi, and P. Frossard, "Deepfool: a simple and accurate method to fool deep neural networks," in *Proceedings of the IEEE Conference on Computer Vision and Pattern Recognition*, 2016, pp. 2574–2582.

[44] F. Croce and M. Hein, "Reliable evaluation of adversarial robustness with an ensemble of diverse parameter-free attacks," in *ICML*, 2020.

[45] F. Tramèr, F. Zhang, A. Juels, M. K. Reiter, and T. Ristenpart, "Stealing machine learning models via prediction apis," in *25th {USENIX} Security Symposium ({USENIX} Security 16)*, 2016, pp. 601–618.

[46] R. Shokri, M. Stronati, C. Song, and V. Shmatikov, "Membership inference attacks against machine learning models," in *2017 IEEE Symposium on Security and Privacy (SP)*. IEEE, 2017, pp. 3–18.

[47] L. Bieringer, K. Grosse, M. Backes, B. Biggio, and K. Krombholz, "Industrial practitioners' mental models of adversarial machine learning," in *Eighteenth Symposium on Usable Privacy and Security (SOUPS 2022)*, 2022, pp. 97–116.

[48] N. Carlini, M. Jagielski, C. A. Choquette-Choo, D. Paleka, W. Pearce, H. Anderson, A. Terzis, K. Thomas, and F. Tramèr, "Poisoning web-scale training datasets is practical," *arXiv preprint arXiv:2302.10149*, 2023.

[49] F. Tramèr, P. Dupré, G. Rusak, G. Pellegrino, and D. Boneh, "Adversarial: Perceptual ad blocking meets adversarial machine learning," in *Proceedings of the 2019 ACM SIGSAC Conference on Computer and Communications Security*, 2019, pp. 2005–2021.

[50] P. Saadatpanah, A. Shafahi, and T. Goldstein, "Adversarial attacks on copyright detection systems," in *International Conference on Machine Learning*. PMLR, 2020, pp. 8307–8315.

[51] AudoTag. AudioTag. Audiotag – free music recognition robot, 2009. [Online]. Available: https://audiotag.info/

[52] Youtube. How content id works – youtube help, 2019.

[53] C. Sitawarin, F. Tramèr, and N. Carlini, "Preprocessors matter! realistic decision-based attacks on machine learning systems," *arXiv preprint arXiv:2210.03297*, 2022.

[54] N. Carlini. A Complete List of All (arXiv) Adversarial Example Papers . [Online]. Available: https://nicholas.carlini.com/writing/2019/all-adversarial-example-papers.html

[55] K. Grosse, L. Bieringer, T. R. Besold, B. Biggio, and K. Krombholz, "Machine learning security in industry: A quantitative survey," *IEEE Transactions on Information Forensics and Security*, vol. 18, pp. 1749–1762, 2023.

[56] Y. Mirsky, A. Demontis, J. Kotak, R. Shankar, D. Gelei, L. Yang, X. Zhang, M. Pintor, W. Lee, Y. Elovici *et al.*, "The threat of offensive AI to organizations," *Computers & Security*, p. 103006, 2022.

[57] N. Carlini. A crisis in adversarial machine learning. [Online]. Available: https://nicholas.carlini.com/talks

[58] O. Bryniarski, N. Hingun, P. Pachuca, V. Wang, and N. Carlini, "Evading adversarial example detection defenses with orthogonal projected gradient descent," *arXiv preprint arXiv:2106.15023*, 2021.

[59] P.-Y. Chen, "Model reprogramming: Resource-efficient cross-domain machine learning," *arXiv preprint arXiv:2202.10629*, 2022.

[60] I. Melnyk, V. Chenthamarakshan, P.-Y. Chen, P. Das, A. Dhurandhar, I. Padhi, and D. Das, "Reprogramming pretrained language models for antibody sequence infilling," 2023.

[61] Y.-Y. Tsai, P.-Y. Chen, and T.-Y. Ho, "Transfer learning without knowing: Reprogramming black-box machine learning models with scarce data and limited resources," in *International Conference on Machine Learning*. PMLR, 2020, pp. 9614–9624.

[62] C.-H. H. Yang, Y.-Y. Tsai, and P.-Y. Chen, "Voice2series: Reprogramming acoustic models for time series classification," in *International Conference on Machine Learning*. PMLR, 2021, pp. 11 808–11 819.

[63] H. A. Dau, A. Bagnall, K. Kamgar, C.-C. M. Yeh, Y. Zhu, S. Gharghabi, C. A. Ratanamahatana, and E. Keogh, "The ucr time series archive," *IEEE/CAA Journal of Automatica Sinica*, vol. 6, no. 6, pp. 1293–1305, 2019.

[64] S. C. Hoffman, V. Chenthamarakshan, K. Wadhawan, P.-Y. Chen, and P. Das, "Optimizing molecules using efficient queries from property evaluations," *Nature Machine Intelligence*, vol. 4, no. 1, pp. 21–31, 2022.

[65] M. T. Ribeiro, S. Singh, and C. Guestrin, "" why should I trust you?" explaining the predictions of any classifier," in *Proceedings of the 22nd ACM SIGKDD International Conference on Knowledge Discovery and Data Mining*, 2016, pp. 1135–1144.

[66] S. M. Lundberg and S.-I. Lee, "A unified approach to interpreting model predictions," in *Proceedings of the 31st International Conference on Neural Information Processing Systems*, 2017, pp. 4768–4777.

[67] H. Ali, M. S. Khan, A. Al-Fuqaha, and J. Qadir, "Tamp-x: Attacking explainable natural language classifiers through tampered activations," *Computers & Security*, vol. 120, p. 102791, 2022.

[68] M. Sundararajan, A. Taly, and Q. Yan, "Axiomatic attribution for deep networks," in *International Conference on Machine Learning*. PMLR, 2017, pp. 3319–3328.

[69] K. Rasheed, A. Qayyum, M. Ghaly, A. Al-Fuqaha, A. Razi, and J. Qadir, "Explainable, trustworthy, and ethical machine learning for healthcare: A survey," *Computers in Biology and Medicine*, p. 106043, 2022.

[70] S. Wachter, B. Mittelstadt, and C. Russell, "Counterfactual explanations without opening the black box: Automated decisions and the gdpr," *Harv. JL & Tech.*, vol. 31, p. 841, 2017.

[71] S. Verma, V. Boonsanong, M. Hoang, K. E. Hines, J. P. Dickerson, and C. Shah, "Counterfactual explanations and algorithmic recourses for machine learning: A review," *arXiv preprint arXiv:2010.10596*, 2020.

[72] A. Dhurandhar, P.-Y. Chen, R. Luss, C.-C. Tu, P. Ting, K. Shanmugam, and P. Das, "Explanations based on the missing: Towards contrastive explanations with pertinent negatives," *Advances in neural information processing systems*, vol. 31, 2018.

[73] B. Ustun, A. Spangher, and Y. Liu, "Actionable recourse in linear classification," in *Proceedings of the Conference on Fairness, Accountability, and Transparency*, 2019, pp. 10–19.

[74] M. Pawelczyk, K. Broelemann, and G. Kasneci, "Learning model-agnostic counterfactual explanations for tabular data," in *Proceedings of The Web Conference 2020*, 2020, pp. 3126–3132.

[75] M. Pawelczyk, C. Agarwal, S. Joshi, S. Upadhyay, and H. Lakkaraju, "Exploring counterfactual explanations through the lens of adversarial examples: A theoretical and empirical analysis," in *International Conference on Artificial Intelligence and Statistics*. PMLR, 2022, pp. 4574–4594.

[76] S. Shan, E. Wenger, J. Zhang, H. Li, H. Zheng, and B. Y. Zhao, "Fawkes: Protecting privacy against unauthorized deep learning models," in *29th USENIX security symposium (USENIX Security 20)*, 2020, pp. 1589–1604.

[77] V. Cherepanova, M. Goldblum, H. Foley, S. Duan, J. Dickerson, G. Taylor, and T. Goldstein, "Lowkey: Leveraging adversarial attacks to protect social media users from facial recognition," *arXiv preprint arXiv:2101.07922*, 2021.

[78] Y. Liang, C. Chen, T. Tian, and K. Shu, "Fair classification via domain adaptation: A dual adversarial learning approach," *Frontiers in Big Data*, vol. 5, p. 129, 2023.

[79] V. Belavadi, Y. Zhou, M. Kantarcioglu, and B. Thuriasingham, "Attacking machine learning models for social good," in *Decision and Game Theory for Security: 11th International Conference, GameSec 2020, College Park, MD, USA, October 28–30, 2020, Proceedings*. Springer, 2020, pp. 457–471.

[80] M. Sharif, S. Bhagavatula, L. Bauer, and M. K. Reiter, "Accessorize to a crime: Real and stealthy attacks on state-of-the-art face recognition," in *Proceedings of the 2016 ACM Sigsac Conference on Computer and Communications Security*, 2016, pp. 1528–1540.

[81] J. Zhang, L. Zhang, G. Li, and C. Wu, "Adversarial examples for good: Adversarial examples guided imbalanced learning," in *2022 IEEE International Conference on Image Processing (ICIP)*. IEEE, 2022, pp. 136–140.

[82] J. Li, T. Zhu, W. Ren, and K.-K. Raymond, "Improve individual fairness in federated learning via adversarial training," *Computers & Security*, p. 103336, 2023.

[83] S. Wang, X. Wang, P.-Y. Chen, P. Zhao, and X. Lin, "Characteristic examples: High-robustness, low-transferability fingerprinting of neural networks." in *IJCAI*, 2021, pp. 575–582.

[84] O. Aramoon, P.-Y. Chen, and G. Qu, "Don't forget to sign the gradients!" *Proceedings of Machine Learning and Systems*, vol. 3, pp. 194–207, 2021.

[85] S. Santurkar, A. Ilyas, D. Tsipras, L. Engstrom, B. Tran, and A. Madry, "Image synthesis with a single (robust) classifier," *Advances in Neural Information Processing Systems*, vol. 32, 2019.

[86] H. Salman, A. Ilyas, L. Engstrom, A. Kapoor, and A. Madry, "Do adversarially robust imagenet models transfer better?" *Advances in Neural Information Processing Systems*, vol. 33, pp. 3533–3545, 2020.

[87] C.-Y. Hsu, P.-Y. Chen, S. Lu, S. Liu, and C.-M. Yu, "Adversarial examples can be effective data augmentation for unsupervised machine learning," in *AAAI Conference on Artificial Intelligence*, 2022.

[88] M. Osadchy, J. Hernandez-Castro, S. Gibson, O. Dunkelman, and D. Peŕez-Cabo, "No bot expects the deepcaptcha! introducing immutable adversarial examples, with applications to captcha generation," *IEEE Transactions on Information Forensics and Security*, vol. 12, no. 11, pp. 2640–2653, 2017.

[89] D. Hitaj, B. Hitaj, S. Jajodia, and L. V. Mancini, "Capture the bot: Using adversarial examples to improve captcha robustness to bot attacks," *IEEE Intelligent Systems*, vol. 36, no. 5, pp. 104–112, 2020.

[90] R. Shao, Z. Shi, J. Yi, P.-Y. Chen, and C.-J. Hsieh, "Robust text captchas using adversarial examples," *arXiv preprint arXiv:2101.02483*, 2021.

[91] M. Gruppi, P.-Y. Chen, and S. Adali, "Fake it till you make it: Self-supervised semantic shifts for monolingual word embedding tasks," in *Proceedings of the AAAI Conference on Artificial Intelligence*, vol. 35, no. 14, 2021, pp. 12 893–12 901.

[92] X. Li, L. Chen, and D. Wu, "Adversary for social good: Leveraging adversarial attacks to protect personal attribute privacy," *arXiv preprint arXiv:2306.02488*, 2023.

[93] F. Brunton and H. Nissenbaum, *Obfuscation: A user's guide for privacy and protest*. Mit Press, 2015.

[94] T. Cilloni, W. Wang, C. Walter, and C. Fleming, "Ulixes: Facial recognition privacy with adversarial machine learning," *Proceedings on Privacy Enhancing Technologies*, vol. 1, pp. 148–165, 2022.

[95] C. Yang, L. Ding, Y. Chen, and H. Li, "Defending against gan-based deepfake attacks via transformation-aware adversarial faces," in *2021 International Joint Conference on Neural Networks (IJCNN)*. IEEE, 2021, pp. 1–8.

[96] R. Wang, Z. Huang, Z. Chen, L. Liu, J. Chen, and L. Wang, "Anti-forgery: Towards a stealthy and robust deepfake disruption attack via adversarial perceptual-aware perturbations," *arXiv preprint arXiv:2206.00477*, 2022.

[97] Z. He, W. Wang, W. Guan, J. Dong, and T. Tan, "Defeating deepfakes via adversarial visual reconstruction," in *Proceedings of the 30th ACM International Conference on Multimedia*, 2022, pp. 2464–2472.

[98] S. Al-Maliki, M. Abdallah, J. Qadir, and A. Al-Fuqaha, "Defending emotional privacy with adversarial machine learning for social good," 2023.

[99] A. Sablayrolles, M. Douze, C. Schmid, and H. Jéǵou, "Radioactive data: tracing through training," in *International Conference on Machine Learning*. PMLR, 2020, pp. 8326–8335.

[100] F. Cayre, C. Fontaine, and T. Furon, "Watermarking security: theory and practice," *IEEE Transactions on signal processing*, vol. 53, no. 10, pp. 3976–3987, 2005.

[101] J. Lee, E. Kim, and S. Yoon, "Anti-adversarially manipulated attributions for weakly and semi-supervised semantic segmentation," in *Proceedings of the IEEE/CVF Conference on Computer Vision and Pattern Recognition*, 2021, pp. 4071–4080.

[102] C. Shi, X. Xu, S. Ji, K. Bu, J. Chen, R. Beyah, and T. Wang, "Adversarial captchas," *IEEE transactions on cybernetics*, 2021.

[103] J. Zhang, J. Sang, K. Xu, S. Wu, X. Zhao, Y. Sun, Y. Hu, and J. Yu, "Robust captchas towards malicious ocr," *IEEE Transactions on Multimedia*, vol. 23, pp. 2575–2587, 2020.

[104] M. Z. Hameed, A. Gÿorgy, and D. Gündüz, "The best defense is a good offense: Adversarial attacks to avoid modulation detection," *IEEE Transactions on Information Forensics and Security*, vol. 16, pp. 1074–1087, 2020.

[105] H. Salman, A. Ilyas, L. Engstrom, S. Vemprala, A. Madry, and A. Kapoor, "Unadversarial examples: Designing objects for robust vision," *Advances in Neural Information Processing Systems*, vol. 34, pp. 15 270–15 284, 2021.

[106] S. Cresci, M. Petrocchi, A. Spognardi, and S. Tognazzi, "Adversarial machine learning for protecting against online manipulation," *IEEE Internet Computing*, vol. 26, no. 2, pp. 47–52, 2021.

[107] S. Al-Maliki, F. E. Bouanani, M. Abdallah, J. Qadir, and A. Al-Fuqaha, "Addressing data distribution shifts in online machine learning powered smart city applications using augmented test-time adaptation," *arXiv preprint arXiv:2211.01315*, 2023.

[108] S. Al-Maliki, F. El Bouanani, K. Ahmad, M. Abdallah, D. T. Hoang, D. Niyato, and A. Al-Fuqaha, "Toward improved reliability of deep learning based systems through online relabeling of potential adversarial attacks," *IEEE Transactions on Reliability*, 2023.

[109] B. Consortium *et al.*, "British national corpus," *Oxford Text Archive Core Collection*, 2007.

[110] M. Davies, "The 385+ million word corpus of contemporary american english (1990–2008+): Design, architecture, and linguistic insights," *International journal of corpus linguistics*, vol. 14, no. 2, pp. 159–190, 2009.

[111] Y. Bengio, A. Cohen, B. Prud'Homme, A. L. D. L. Alves, and o. Oder, "Innovation ecosystems for socially beneficial AI," *Missing links in AI governance*, p. 133, 2023.

[112] E. Radiya-Dixit, S. Hong, N. Carlini, and F. Trame`r, "Data poisoning won't save you from facial recognition," *arXiv preprint arXiv:2106.14851*, 2021.

[113] Y. Zheng, X. Feng, Z. Xia, X. Jiang, M. Pintor, A. Demontis, B. Biggio, and F. Roli, "Stateful detection of adversarial reprogramming," *Information Sciences*, p. 119093, 2023.

[114] S. Dai, S. Mahloujifar, C. Xiang, V. Sehwag, P.-Y. Chen, and P. Mittal, "Multirobustbench: Benchmarking robustness against multiple attacks," 2023.

[115] HAI. Artificial Intelligence Index Report 2023. [Online]. Available: https://aiindex.stanford.edu/wp-content/uploads/2023/04/HAI_AI-Index-Report_2023.pdf

[116] G. of high-profile researchers and business leaders. Pause Giant AI Experiments: An Open Letter. [Online]. Available: https://futureoflife.org/open-letter/pause-giant-ai-experiments/

[117] UN. UN Guiding Principles on Business and Human Rights. [Online]. Available: https://www.business-humanrights.org/en/